\documentclass[10pt]{article} 
\usepackage[accepted]{tmlr}

\usepackage{amsmath,amsfonts,bm}

\def\eqref#1{equation~\ref{#1}}

\def\1{\bm{1}}

\DeclareMathAlphabet{\mathsfit}{\encodingdefault}{\sfdefault}{m}{sl}
\SetMathAlphabet{\mathsfit}{bold}{\encodingdefault}{\sfdefault}{bx}{n}

\usepackage{hyperref}
\usepackage{url}
\usepackage{microtype}
\usepackage{graphicx}
\usepackage{booktabs} 

\usepackage{amsmath}
\usepackage{amssymb}
\usepackage{mathtools}
\usepackage{amsthm}
\usepackage{caption}
\usepackage{subcaption}
\usepackage{threeparttable}
\usepackage{multirow}
\usepackage{enumitem}
\usepackage[capitalize,noabbrev]{cleveref}

\usepackage[ruled, vlined, linesnumbered, algo2e]{algorithm2e}%
\usepackage{wrapfig}
\usepackage{algorithm}
\usepackage{algorithmic}

\AtBeginEnvironment{algorithmic}{}

\theoremstyle{plain}

\theoremstyle{definition}

\theoremstyle{remark}

\title{Efficient and Robust Quantization-aware Training \\via Adaptive Coreset Selection}

\author{\name Xijie Huang$^1$, Zechun Liu$^2$, Shih-yang Liu$^1$, Tim Kwang-Ting CHENG$^1$ \\
      \addr  $^1$Hong Kong University of Science and Technology (HKUST), $^2$Meta Reality Lab \\
      xhuangbs@connect.ust.hk, zechunliu@meta.com, sliuau@connect.ust.hk, timcheng@ust.hk }

\def\month{08}  
\def\year{2024} 
\def\openreview{\url{https://openreview.net/forum?id=4c2pZzG94y}} 

\begin{document}

\maketitle

\begin{abstract}
Quantization-aware training (QAT) is a representative model compression method to reduce redundancy in weights and activations. 
However, most existing QAT methods require end-to-end training on the entire dataset, which suffers from long training time and high energy costs. 
In addition, the potential label noise in the training data undermines the robustness of QAT.
In this work, we show that we can improve data efficiency and robustness with a proper data selection strategy designed specifically for QAT.
We propose two metrics based on analysis of loss and gradient of quantized weights: error vector score and disagreement score, to quantify the importance of each sample during training.
Guided by these two metrics, we proposed a quantization-aware \textbf{\underline{A}}daptive \textbf{\underline{C}}oreset \textbf{\underline{S}}election (\textbf{ACS}) method to select the data for the current training epoch. 
We evaluate our method on various networks (ResNet-18, MobileNetV2, RetinaNet), datasets(CIFAR-10, CIFAR-100, ImageNet-1K, COCO), and under different quantization settings.
Specifically, our method can achieve an accuracy of 68.39\% of 4-bit quantized ResNet-18 on the ImageNet-1K dataset with only a 10\% subset, which has an absolute gain of 4.24\% compared to the baseline. Our method can also improve the robustness of QAT by removing noisy samples in the training set. 
\end{abstract}

\section{Introduction}
\begin{figure}[h]
  \centering
    \includegraphics[width=0.36\textwidth]{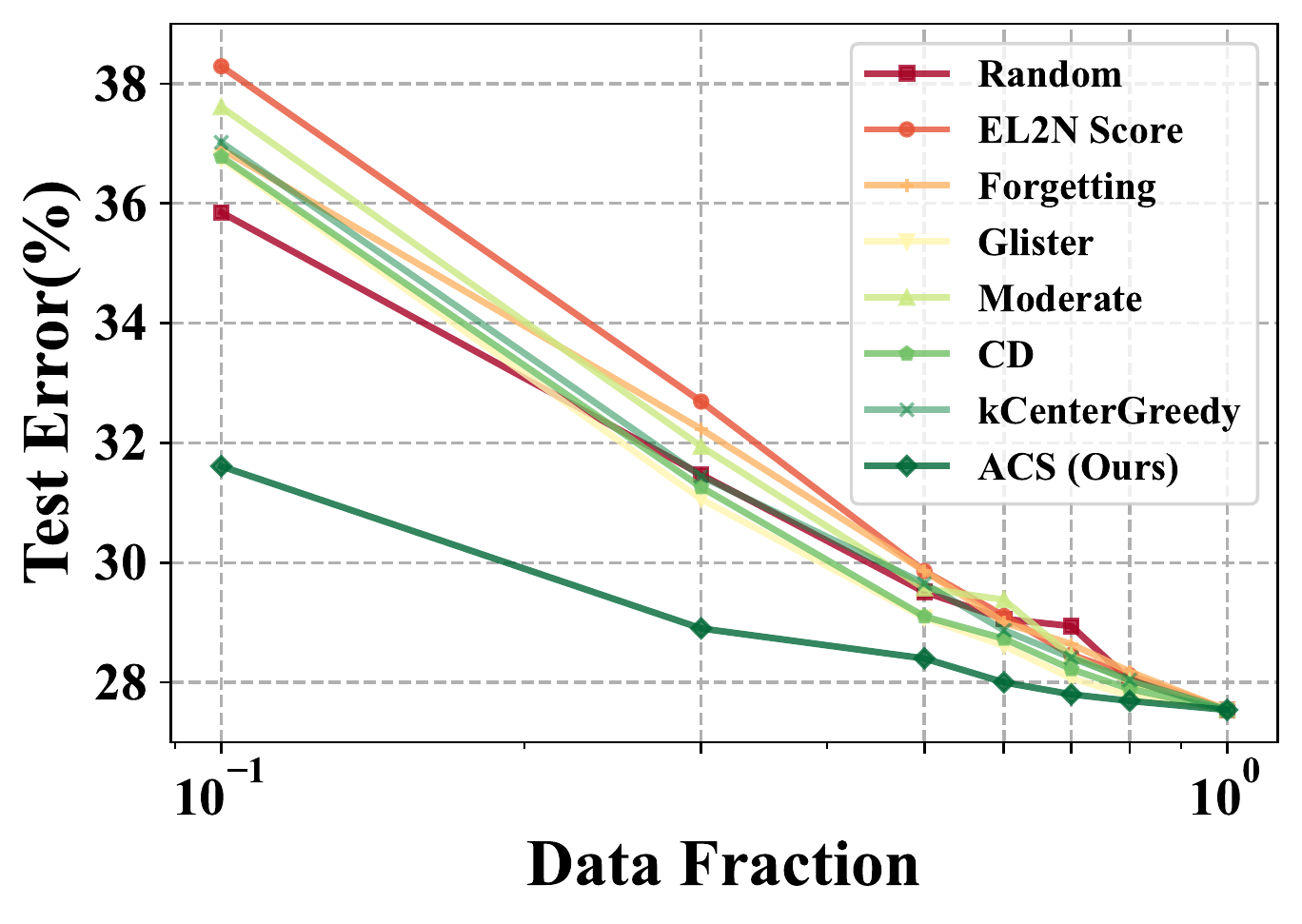}
    \includegraphics[width=0.36\textwidth]{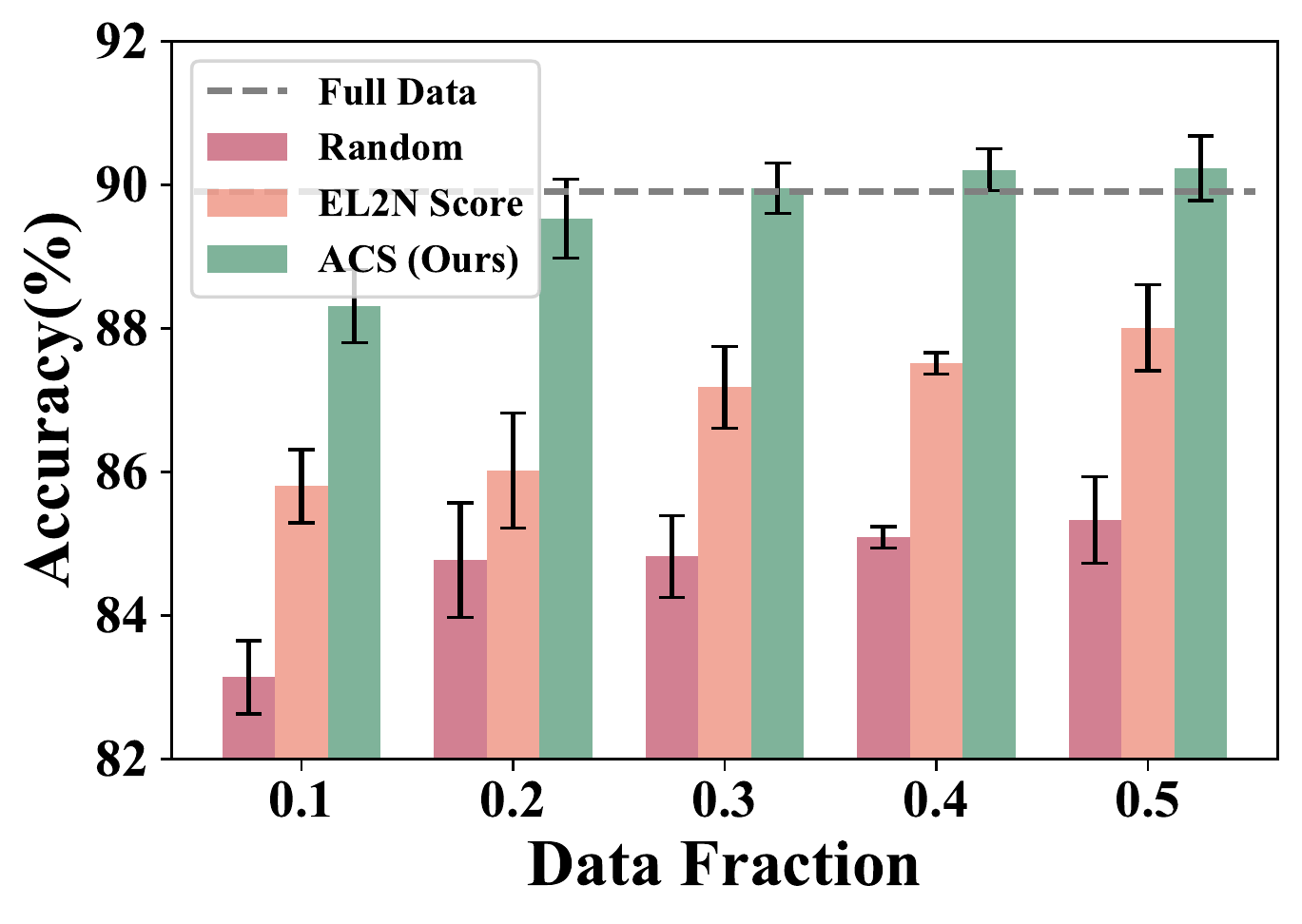}
    \includegraphics[width=0.26\textwidth]{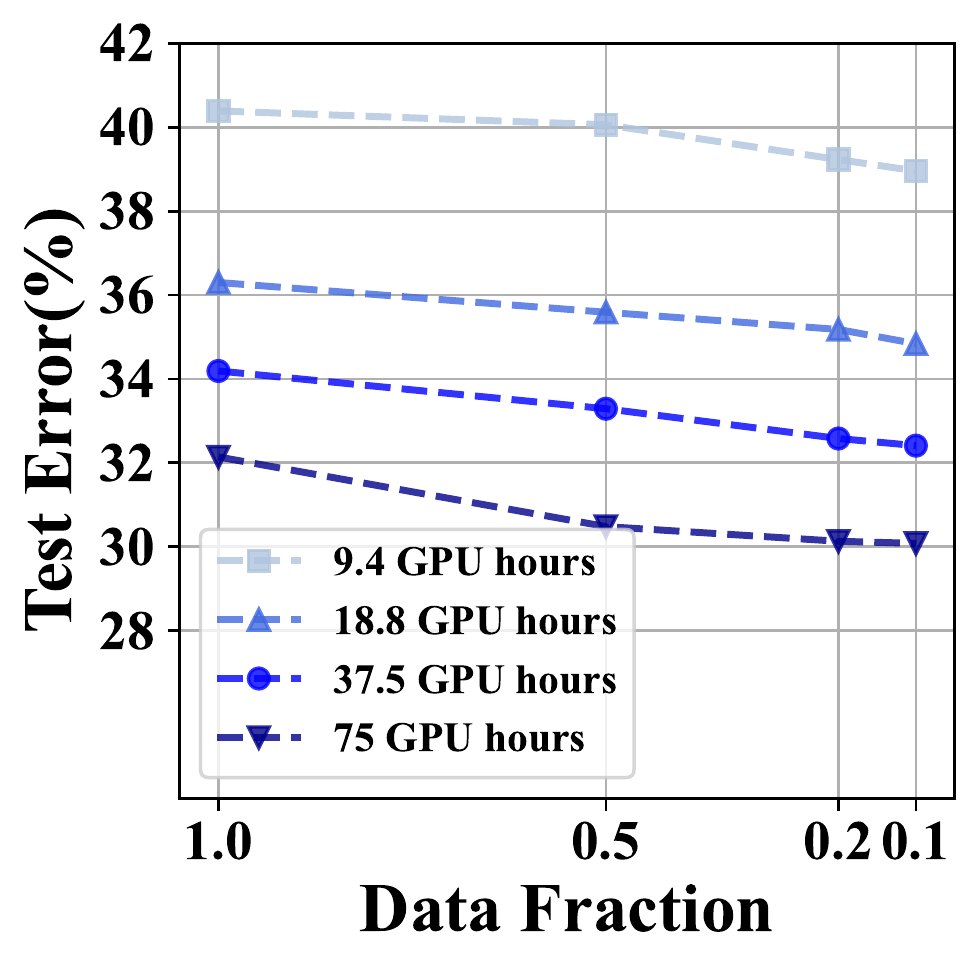}
    \vspace{-0.1in}
  \caption{\textbf{Left:} Data scaling curve for 4-bit quantized ResNet-18 on the ImageNet-1K dataset. Our ACS significantly reduces test error using the same training data fraction compared to baselines. \textbf{Middle:} Accuracy of 2/32-bit quantized ResNet-18 trained on CIFAR-10 with 10\% random label noise. Our ACS is the only method to outperform the full data training performance by effectively removing noisy samples. \textbf{Right:} Test error for quantized ResNet-18 with different data fraction and the same GPU training hours. Under the same training budget, QAT with smaller coreset selected by ACS outperforms full data training.} \label{fig:scaling_curve}
\end{figure}

\begin{figure*}[t]
  \centering
    \includegraphics[width=0.95\textwidth]{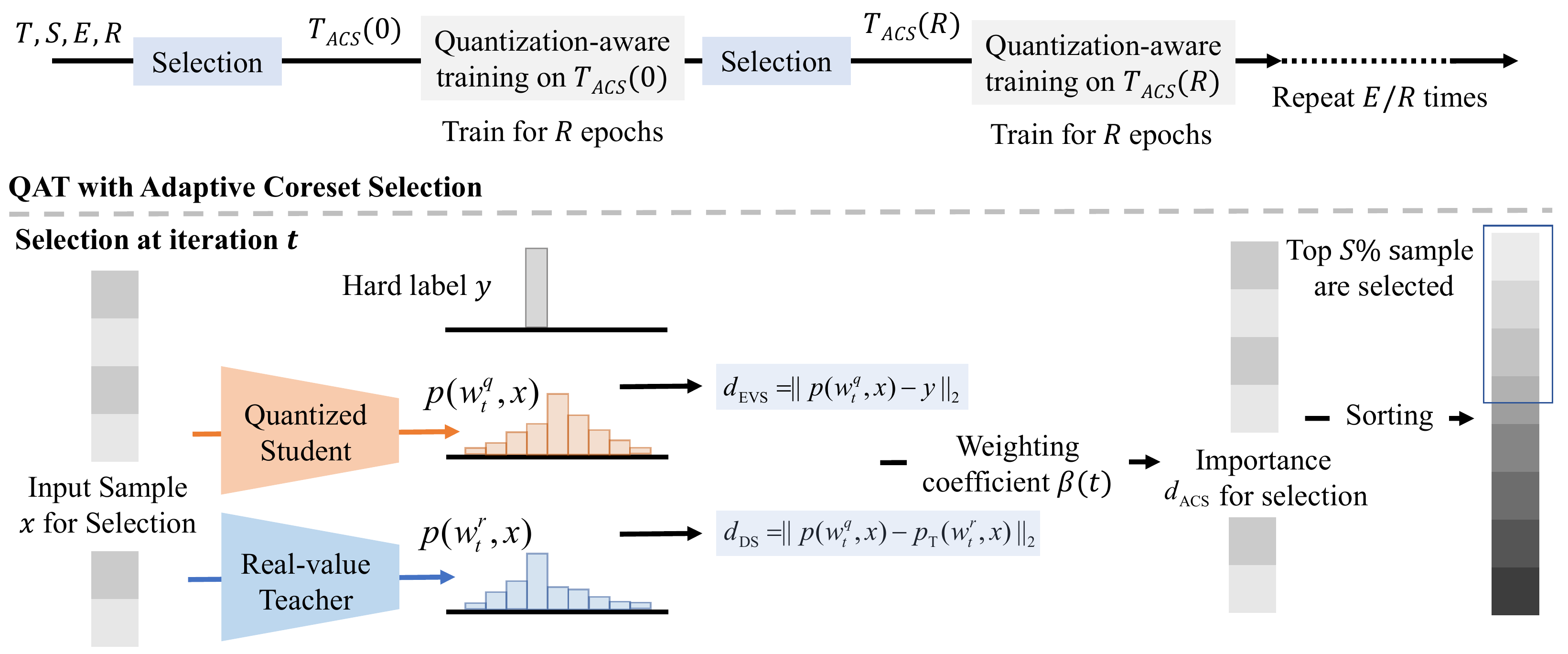}
    \caption{An overview of the Adaptive Coreset Selection (ACS) for efficient and robust QAT. }
    \label{fig:acs}
\end{figure*}

Deep learning models have achieved remarkable achievements across various applications, including computer vision~\citep{krizhevsky2012imagenet, he2016deep, tan2019efficientnet, kirillov2023segment} and natural language processing~\citep{kenton2019bert, yang2019xlnet, conneau2019cross, touvron2023llama}. 
The outstanding performance of these models can be attributed to their large number of parameters and the availability of large-scale training data. For example, GPT-3~\citep{brown2020language} boasts 175 billion parameters and the pre-training is performed on the dataset with more than 410 billion tokens. Neural scaling law~\cite{kaplan2020scaling,alabdulmohsin2022revisiting} is an explanation of this size expansion that empirically demonstrates test loss of a model follows a power law to the model size, training dataset size, and amount of computation. While neural scaling law guides the expanding usage of large models, the large model size and training set scale have become the most significant challenges for the training and deployment of deep learning models, especially on edge devices with computation and storage constraints. 

Many model compression methods have been proposed recently to address these challenges and enable the effective deployment of deep learning models. These model compression techniques include quantization~\citep{zhou2016dorefa,choi2018pact,esser2020learned,bhalgat2020lsq+}, pruning~\citep{liu2017learning, liu2018rethinking,molchanov2019importance,liu2019metapruning}, knowledge distillation~\citep{hinton2015distilling,park2019relational,shen2021fast}, and compact network design~\citep{howard2017mobilenets, pham2018efficient}. Among the aforementioned methods, quantization methods have been the most widely adopted because they have the advantage of promising hardware affinity across different architectures~\citep{judd2016stripes,jouppi2017datacenter,sharma2018bit}. To minimize the performance gap between the quantized and full-precision models, quantization-aware training (QAT) is often utilized. Although QAT has improved the inference efficiency of the target model, it is computation-intensive and requires more time than full-precision training. While previous QAT methods assume an ideal case in which computation resources are unlimited, we aim to research a more practical case when the cost of QAT should be considered.

Coreset selection techniques aim to mitigate the high training cost and potential negative influence of label noise to improve data efficiency and robustness for full-precision training. Specifically, coreset selection methods leverage the redundancy in training datasets and select the most informative data to build a coreset for training. One recent research~\citet{sorscher2022beyond} has pointed out that coreset can be a solution to beat the scaling law.Previous methods select data based on feature~\citep{agarwal2020contextual, seneractive}, error~\citep{paul2021el2n}, gradient~\citep{killamsetty2021grad}, and decision boundary~\citep{margatina2021active}. While these methods can achieve notable efficiency improvement for full-precision training, their effectiveness on low-precision QAT has not been explored before.
To utilize coreset selection methods to improve the data efficiency and robustness of QAT, the characteristics of quantized weights must be considered in the design of methods. Existing full-precision methods are not designed for quantization, and severe computation overhead during selection may hinder the application of QAT. Therefore, a coreset selection method specifically designed for QAT is required.

In this work, we propose the coreset selection specifically for QAT and find that we can elevate the data efficiency and training robustness.
We start by analyzing the impact of removing one specific sample from the training set during QAT and identifying that the error vector score (EVS) is a good theoretical approximation of the importance of each sample. Based on the common practice of utilizing knowledge distillation during QAT, we also propose the disagreement score (DS) measuring the prediction gap between the quantized model and full-precision models. Based on these metrics, we propose a fully quantization-aware adaptive coreset selection (ACS) method to select training samples that fit the current training objective. An overview of the proposed method is shown in Fig.~\ref{fig:acs}.

We demonstrate the superiority of our ACS in terms of effectiveness and efficiency on different networks (ResNet-18, MobileNet-V2, RetinaNet), datasets (CIFAR-10, CIFAR-100, ImageNet-1K, COCO), and quantization settings. For 2-bit weights-only quantization of MobileNetV2 on the CIFAR-100 dataset, QAT based on our ACS can achieve a mean accuracy of 67.19\% with only 50\% training data used for training per epoch. For 4-bit quantization of ResNet-18 on the ImageNet-1K dataset, our ACS can achieve top-1 accuracy of 68.39\% compared to the 64.15\% of random selection when only 10\% training data is selected for the training of every epoch.
In summary, our contribution can be summarized as follows:
\begin{itemize}[itemsep=3pt,topsep=0pt,parsep=0pt]
    \item We are \textbf{the first} to investigate the data efficiency and robustness of quantization-aware training and empirically observe that the importance of different data samples varies during the training process. 
    
    \item We propose two metrics: error vector score and disagreement score, to quantify the importance of each sample based on theoretical analysis of loss gradient.
    
    \item We propose a \textbf{quantization-aware} \textbf{\underline{A}}daptive \textbf{\underline{C}}oreset \textbf{\underline{S}}election (\textbf{ACS}) method, which adaptively selects informative samples that fit current training epochs and prune redundant or noisy samples.
    
    \item We verify ACS on different network architectures, datasets, and quantization settings. Compared with previous methods, ACS can significantly improve both the data efficiency and robustness of quantization-aware training shown in Fig.~\ref{fig:scaling_curve}.
\end{itemize}

\section{Related Work}


\paragraph{Quantization} 
Quantization methods are powerful tools for improving the efficiency of model inference. The core insight is replacing full-precision weights and activations with lower-precision representation. The quantization methods can be classified as quantization-aware training (QAT)~\citep{zhou2016dorefa,esser2020learned,bhalgat2020lsq+,huang2022sdq,liu2023llmqat} and post-training quantization (PTQ)~\citep{nagel2020up,fang2020post,wang2020towards,xiao2023smoothquant,liu2023llm,FlexRound} based on whether to retrain a model with quantized weights and activations or start with a pre-trained model and directly quantize it without extensive training. 
Based on the characteristics of quantization intervals, these methods can be categorized into uniform and non-uniform quantization. While uniform quantization~\citep{zhou2016dorefa,choi2018pact,esser2020learned} with uniform interval are more hardware-friendly and efficient, non-uniform quantization~\citep{miyashita2016convolutional, zhang2018lqnet, li2019apot,yvinec2023powerquant}, due to the flexibility of representation, can minimize the quantization error and achieve better performance than uniform schemes. While previous QAT methods assume an ideal case for unlimited computation resources, we research a more practical case to conduct QAT with a fixed computation budget in this work.

\paragraph{Learning from Noisy Labels} The \textit{noisy labels} are defined as unreliable labels that are corrupted from ground truth. The noise can stem from non-expert labeling or malicious label-flipping attack~\citep{xiao2012adversarial}. The estimated ratio of noisy labels in real-world datasets ranges from 8.0\%-38.5\% according to \cite{song2022learning}. To improve the robustness of learning for noisy labels, the solutions include designing robust architectures that are attuned to label noise~\citep{yao2018deep,lee2019robust}, label correction~\citep{tu2023learning,li2023disc,li2024nicest}, robust regularization~\citep{wei2021open}, and robust loss design~\citep{huang2023twin}. One important category that enables robust noise-tolerant training is sample selection~\citep{paul2021el2n,xia2023combating,park2024robust}. While the effectiveness of sample selection methods has been verified on full-precision deep learning, their performance on QAT is under-explored.

\paragraph{Coreset Selection}
Coreset selection targets improving the data efficiency by identifying the informative training samples. Previous works can be mainly classified into the following categories: 
\textbf{Geometry-based Methods:} These methods assume that samples close to each other in the feature space similarly influence the training. These redundant data points should be removed to improve efficiency. Some representative works include Contextual diversity (CD)~\citep{agarwal2020contextual}, k-Center-Greedy~\citep{seneractive}, ~\citet{sorscher2022beyond}, Moderate Coreset~\citep{xia2023moderate}, and Herding~\citep{welling2009herding}.
\textbf{Decision Boundary-based Methods:} These methods select samples distributed near the decision boundary. These samples are difficult for the model to separate. Representative works include Adversarial Deepfool~\citep{ducoffe2018adversarial} and Contrastive Active Learning (CAL)~\citep{margatina2021active}. 
\textbf{Gradient/Error-based Methods:} These methods assume the samples are more informative if they contribute more to the error or loss during the training. This category includes EL2N~\citep{paul2021el2n},  CRAIG~\citep{mirzasoleiman2020coresets}, GradMatch~\citep{killamsetty2021grad}, Forgetting events~\citep{toneva2018forgetting}, and AdaCore~\citep{pooladzandi2022adaptive}.
\textbf{Optimization-based Methods:} These methods formulate the coreset selection as a bilevel optimization problem. The outer level objective is the selection of samples and the inner level objective is the optimization of model parameters. The representative works include~\citet{borsos2020coresets}, Glister~\citep{killamsetty2021glister},~\citet{zhou2022probabilistic}, and Retrieve~\citep{killamsetty2021retrieve}. 

These methods perform well in some scenarios of full-precision training(specific subset fraction, specific outlier distribution, etc.). However, \textbf{none are verified on quantization settings or consider the requirements for QAT}. As shown in Fig.~\ref{fig:scaling_curve}, the majority of these methods cannot even outperform random sampling in the QAT settings. Moreover, these methods either require early training of the target model for several epochs~\citep{toneva2018forgetting, paul2021el2n} or time-consuming search~\citep{seneractive} and optimization~\citep{killamsetty2021glister,killamsetty2021retrieve}, which leads to heavy computation overhead during QAT.

\section{Importance of Each Sample in QAT} \label{sec:importance}
One underlying assumption of data scaling law is that all samples in the training set are equally important. However, many previous coreset research~\cite{sorscher2022beyond} have proved that information within different samples is different for full-precision training, and the quantization setting has not been verified yet. 
In this section, we will first introduce quantization-aware training (QAT) and derive the gradient under cross-entropy and SGD in Sec.~\ref{sec:qat}. We then analyze the change of gradient when a specific sample is removed from a training batch to investigate the importance of each training sample in Sec.~\ref{sec:evs}. We propose an approximation of this gradient change considering the prediction error without introducing any memory overhead. We further prove that less training data is required when knowledge distillation is applied to QAT in Sec.~\ref{sec:ds}. Another metric quantifying sample importance based on the prediction gap between the quantized student and full-precision teacher model is also introduced in Sec.~\ref{sec:ds}.

\subsection{Preliminaries of QAT} \label{sec:qat}
During QAT, the real-value data $v^r$ is converted to $b$-bit quantized representation $v^q=q_b(v^r)$ by the quantizer $q_b$. Given the scale factor $s$ of the quantizer, the number of positive quantization levels $Q_P$, and the number of negative quantization levels $Q_N$, we can have the quantizer $q_b$ as
\begin{equation}\label{equ:quantizer}
   v^q=q_b(v^r)=s \times \lfloor \text{clip}(v^r/s, -Q_N, Q_P) \rceil,
\end{equation}
where $\lfloor \cdot \rceil$ is the rounding function that rounds the input to the nearest integer, $\text{clip}(v, r_{\text{low}}, r_{\text{high}})$ return $v$ with all value below $r_{\text{low}}$ set to be $r_{\text{low}}$ and all values above $r_{\text{high}}$ set to be $r_{\text{high}}$. For the unsigned quantization, $Q_N=0, Q_P=2^b-1$. While for the quantization of signed data, $Q_N=2^{b-1}, Q_P=2^{b-1}-1$. To solve the problem that the gradient cannot back-propagate in Equation~\ref{equ:quantizer} during QAT, the straight-through estimator (STE)~\citep{bengio2013estimating} is utilized to approximate the gradient. In the back-propagation of QAT with STE, the gradient of the loss $\mathcal{L}$ with respect to the real-value data $v^r$ is set to be
\begin{equation}  \label{equ:STE}
\begin{aligned}
   \frac{\partial \mathcal{L}}{\partial v^r}= \frac{\partial \mathcal{L}}{\partial v^q} \cdot \textbf{1}_{-Q_N \leq v^r/s \leq Q_P},
\end{aligned}
\end{equation}
where $\textbf{1}$ is the indicator function that outputs 1 within the quantization limit and 0 otherwise. Note that as we will apply clipping in the beginning of QAT, the gradient of quantized data can be seen as approximately equal as the gradient of real-value data.

The training set of QAT is denoted as $\mathcal{T} = \{(x_i, y_i)\}_{i=1}^N$, where input $x_i \in \mathbb{R}^d$ are vectors of length $d$ and $y \in \{0,1\}^M$ are one-hot vectors encoding labels. The neural network to be trained is denoted as $f(w^r,x)$, where real-value weights are $w^r$. We use cross-entropy loss $\mathcal{L}(\hat p, y) = \sum_{m=1}^{M} y^{(m)} \log p^{(m)}$ in the QAT and $p(w^q,x)$ is a probability vector denoting the output of the quantized neural network $f(w^q,x)$. Stochastic gradient descent (SGD) is used for the optimization. Suppose the real-value weights at iteration $t$ are $w^r_t$ and the batch of input samples is $\mathcal{T}_{t-1} \subseteq \mathcal{T}$, the weights are updated following 
\begin{equation} \label{eq:discrete}
    w^r_t = w^r_{t-1} - \eta  \sum_{(x,y) \in \mathcal{T}_{t-1}} g_{t-1}(x,y),
\end{equation}
where $\eta$ denotes the learning rate and $g_{t-1}(x,y) = \nabla \mathcal{L}_{t-1}(p(w^r_{t-1},x),y)$ is the gradient of cross entropy loss. 

\subsection{Error Vector Score} \label{sec:evs}

To look into the importance of each sample $\{(x_i, y_i)\}$, we measure the difference of the expected magnitude of the loss vector on the training set $\mathcal{T}$ and another training set which only removes one specific sample $ \mathcal{T}' = \mathcal{T} \setminus \{(x_i, y_i)\}$. 
 For simplicity, we approximate all the training dynamics in continuous time. Based on the chain rule and STE of quantization, we have the change of loss $\mathcal{L}$ at time $t$ on sample $(x,y)$ of batch $\mathcal{T}$ as
 \begin{equation} \small
     \left.\frac{d \mathcal{L}}{d t}\right|_{(x,y),\mathcal{T}_t} = g_t(x,y)\frac{d w^q_t}{d t} = g_t(x,y)\frac{d w^r_t}{d t} .
 \end{equation}
According to the discrete-time dynamic of real-value weights in Eq.~\ref{eq:discrete}, we have $\frac{d w^q_t}{d t} \approx w^r_t - w^r_{t-1} = - \eta  \sum_{(x,y) \in \mathcal{T}_{t-1}} g_{t-1}(x,y)$. To measure the contribution of a specific sample $(x_i, y_i)$, we measure the change of loss with and without the sample. For a given data batch $\mathcal{T}$, if the sample $(x_i, y_i) \notin \mathcal{T}$, we can ignore the change in $\left.\frac{d \mathcal{L}}{d t}\right|_{(x,y),\mathcal{T}_t}$. For any sample $(x_j, y_j) \in \mathcal{T}, j \neq i$ in the same batch, the importance $\mathcal{I}(x_i, y_i)$ is measured as
\begin{equation} \small
\mathcal{I}(x_i, y_i) = \left\| \left.\frac{d \mathcal{L}}{d t}\right|_{(x_j, y_j),\mathcal{T}} - \left.\frac{d \mathcal{L}}{d t}\right|_{(x_j, y_j), \mathcal{T}'} \right\|.
\end{equation}

According to the chain rule, we have 
\begin{equation} \small
\left.\frac{d \mathcal{L}}{d t}\right|_{(x_j, y_j),\mathcal{T}} = \frac{d \mathcal{L}(p(w^q_t,x_j), y_j)}{d w^q_t} \frac{d w^q_t}{d w^r_t} \frac{d w^r_t}{dt}
\end{equation}
According to the characteristics of STE in Eq.~\ref{equ:STE}, $\frac{d w^q_t}{d w^r_t}=1$ holds for all input within the clipping range. Following the updating rule of weights in Eq. ~\ref{eq:discrete}, the gradient $\frac{d w^r_t}{dt} = - \eta  \sum_{(x^*,y^*) \in \mathcal{T}} g_{t-1}(x^*,y^*)$. 
The only difference of the training set $\mathcal{T}$ and $\mathcal{T}'$ is the existence of sample $(x_i,y_i)$. Thus, we have
\begin{equation}\small
\begin{aligned}
\mathcal{I}(x_i, y_i) &= \left\| \frac{d \mathcal{L}}{d w^q_t} (\left.\frac{d w^r_t}{dt}\right|_{(x_j, y_j),\mathcal{T}} - \left.\frac{d w^r_t}{dt}\right|_{(x_j, y_j),\mathcal{T'}}) \right\| 
\\ &= \eta \left\|\frac{d \mathcal{L}}{d w^q_t} \cdot g_{t-1}(x_i,y_i) \right\|.
\end{aligned}
\end{equation}

We use $\frac{d \mathcal{L}}{d w^q_t}$ as a simplification of $\frac{d \mathcal{L}(p(w^q_t,x_j), y_j)}{d w^q_t}$, which is only dependant on the current training sample $(x_j,y_j)$ but not dependant on the sample $(x_i,y_i)$ that is removed from batch $\mathcal{T}$.
Since learning rate $\eta$ and the gradient of loss w.r.t. quantized weights $\frac{d \mathcal{L}}{d w^q_t}$ can be seen as constant given $\mathcal{T}$, the importance of the sample $(x_i,y_i)$ in this batch $\mathcal{T}$ is only related to the gradient norm of cross-entropy loss of this sample $||g_{t-1}(x_i,y_i)||$. The examples with a larger gradient norm expectation have more influence on the supervised training of other data, which means they are important for QAT and should be included in the coreset. We can select data with high importance by sorting by the norm of gradient, which is also covered in previous works~\cite{paul2021el2n}. However, storing the loss gradient for comparison requires extra memory and is hard to transfer between network architectures. We approximate the norm gradient with the norm of the error vector, which is defined as follows.
\paragraph{Definition 1 (Error Vector Score)} \emph{
The error vector score of a training sample $(x,y)$ at iteration $t$ id defined to be $d_{\text{EVS}} = \| p(w^q_t,x) - y \|_2$. 
}

For any input $x \in \mathbb{R}^d$, gradient norm $||g_{t}(x,y)||$ is a non-random score. We take its expectation over random minibatch sequence and random initialization to get the expected gradient norm $\mathbb{E}||g_{t}(x,y)||$, which can also be indicated as
\begin{equation} \small
\mathbb{E}\left\|g_{t}(x,y)\right\| =  \sum_{m=1}^{M} \frac{d \mathcal{L}_{t}(p(w^q_{t},x),y)^{T}}{d f^{(m)}_t} \frac{d f^{(m)}_t}{d w^q_t},
\end{equation}

where $\frac{d f^{(m)}_t}{d w^q_t}$ denotes the gradient of $m$-th logit on weights. Since we are using cross-entropy as the loss function, the gradient of loss $\mathcal{L}$ on the $m$-th logit output $f^{(m)}_t$ follows $\frac{d \mathcal{L}_{t}(p(w^q_{t},x),y)^{T}}{d f^{(m)}_t} = p(w^q_t,x)^{(m)} - y^{(m)}$. 
Previous works~\citep{fort2020deep,fort2019emergent} empirically observe that $\frac{d f^{(m)}_t}{d w^q_t}$ are similar across different logits $m \in M$ and training sample $(x,y)$. Thus, selecting samples based on the error vector score $d_{\text{EVS}}$ will be more likely to identify the samples with a larger gradient norm.
Different from the previous method~\citep{paul2021el2n} also leveraging the error metrics, no early training is required, and we only use the current quantized model prediction $p(w^q_{t},x)$ during QAT.

\subsection{Disagreement Score and Knowledge Distillation} \label{sec:ds}

Intrinsically, a quantized classification network should learn an ideal similar mapping $f$ from input sample $x$ to the output logits $f(w,x)$ as a full-precision network, and the gap between the quantized prediction $p(w^q,x)$ of student model and real-value prediction $p_\mathbf{T}(w^r,x)$ of teacher model $\mathbf{T}$ needs to be minimized. Based on this insight, knowledge distillation (KD) is widely used during QAT with a full-precision model as the teacher, which can also be seen in previous works~\citep{polinomodel, huang2022sdq, mishra2018apprentice, liu2023oscillation}. The loss function is designed to enforce the similarity between the output of the full-precision teacher and the quantized student model as
\begin{equation}\label{eq:kd} \small
\begin{aligned}
\begin{split}
    \mathcal{L}_{KD} = -\frac{1}{N}\sum^M_m\sum_{i=1}^{N} p_\mathbf{T}^{(m)}(w^r,x_i)\log(p^{(m)}(w^q,x_i))
\end{split}
\end{aligned}
\end{equation}
where the KD loss is defined as the cross-entropy between the output distributions $p_\mathbf{T}$ of a full-precision teacher and a quantized student on the same input $x$ but different weights representation $w^r$ and $w^q$. $x_i$ is one of the input samples from the training set. $m$ and $N$ denote the classes and total training sample numbers, respectively. 

Note that this process can be regarded as the distribution calibration for the student network and one-hot label is not involved during QAT. Since the loss function used for knowledge distillation is still cross-entropy, and we still assume we use SGD for the optimization, most conclusions in Sec.~\ref{sec:evs} still hold by replacing the one-hot ground truth label $y$ with full-precision teacher's prediction $p_\mathbf{T}(w^r_t,x)$. Thus, we propose the disagreement score as follows.
\paragraph{Definition 2 (Disagreement Score)} \emph{
The disagreement score of a training sample $(x,y)$ at iteration $t$ is defined to be $d_{\text{DS}} = \| p(w^q_t,x) - p_\mathbf{T}(w^r_t,x) \|_2$.
}

The core difference between error vector score $d_{\text{EVS}}$ and disagreement score $d_{\text{DS}}$ is the target label. While $d_{\text{EVS}}$ uses one-hot hard labels, the $d_{\text{DS}}$ uses the distilled soft labels. We empirically notice that the data needed for the training is reduced when knowledge distillation is applied, which is helpful for our selection with a small data fraction. We further demonstrate the advantage in terms of training data requirements using the soft label based on Vapnik–Chervonenkis theory~\citep{vapnik1999overview}, which decomposes the classification error $R(f_{s})$ of a classifier $f_s \in \mathcal{F}_s$ as 
\begin{equation} \small
      R(f_{s}) - R(f) \leq O\left(\frac{|\mathcal{F}_s|_\textrm{C}}{n^{\alpha_s}}\right) + \varepsilon_s,
\end{equation}
where $O(\cdot)$ denotes the asymptotic approximation and $\varepsilon_s$ is the
approximation error of $\mathcal{F}_s$ with respect to $\mathcal{F}$. 
$f \in  \mathcal{F}$ denotes the real target function.
$|\cdot|_\textrm{C}$ is the VC-Dimension of the function class measuring its capacity. $n$ is the number of total training data. $\frac{1}{2} \leq \alpha_s \leq 1$ is an indicator measuring the difficulty of the problems. For non-separable and difficult problems, $ \alpha_s = \frac{1}{2}$, which means the classifier learns at a slow rate of $O(n^{-\frac{1}{2}})$. For separable and easy problems, $ \alpha_s = 1$, indicating the classifier learns at a fast rate. In our setting, if the quantized model $f_q \in  \mathcal{F}_q$ directly learns from the hard labels, the difficulty of the problem is high, and we assume $\alpha_q = \frac{1}{2}$, we have 
\begin{equation} \label{eq:vc} \small
      R(f_{q}) - R(f) \leq O\left(\frac{|\mathcal{F}_q|_\textrm{C}}{\sqrt{n}}\right) + \varepsilon_q,
\end{equation}
where $\varepsilon_q$ is the approximation error of the quantized model. However, if we first train the full-precision teacher model $f_r \in \mathcal{F}_r$ and then utilize knowledge distillation to learn the representation from the teacher, the difficulty of the learning becomes easier,  assuming $\alpha_r = 1$, we have
\begin{equation} \small
\begin{aligned}
      R(f_{r}) - R(f) \leq O\left(\frac{|\mathcal{F}_r|_\textrm{C}}{n}\right) + \varepsilon_r, && R(f_{q}) - R(f_r) \leq O\left(\frac{|\mathcal{F}_q|_\textrm{C}}{n^{\alpha_{qr}}}\right) + \varepsilon_{qr}, 
\end{aligned}
\end{equation}
where the $\varepsilon_r, \varepsilon_{qr}$ denotes the approximation error of the $\mathcal{F}_r$ with respect to $\mathcal{F}$ and approximation error of the $\mathcal{F}_q$ with respect to $\mathcal{F}_r$, respectively. Compared to the direct learning quantized model from the hard label shown in Eq.~\ref{eq:vc}, the knowledge distillation with real-value teacher $f_r$ yields the classification error as follows:
\begin{equation} \small
\begin{aligned}
    R(f_{q}) - R(f) 
    \leq O\left(\frac{|\mathcal{F}_r|_\textrm{C}}{n}\right) + \varepsilon_r + O\left(\frac{|\mathcal{F}_q|_\textrm{C}}{n^{\alpha_{qr}}}\right) + \varepsilon_{qr} \leq O\left(\frac{|\mathcal{F}_q|_\textrm{C} + |\mathcal{F}_r|_\textrm{C}}{n^{\alpha_{qr}}}\right) + \varepsilon_r+\varepsilon_{qr}.
\end{aligned}
\end{equation}
Following previous studies on knowledge distillation~\citep{lopez2015unifying,mirzadeh2020improved}, the soft labels contain more information than hard labels for each sample. Thus, we have $\varepsilon_r+\varepsilon_{qr} \leq \varepsilon_q$ and $O\left(\frac{|\mathcal{F}_q|_\textrm{C} + |\mathcal{F}_r|_\textrm{C}} {n^{\alpha_{qr}}}\right) \leq O\left(\frac{|\mathcal{F}_q|_\textrm{C}}{\sqrt{n}}\right)$. Combining these two inequalities, we have the inequality
\begin{equation} \small
    O\left(\frac{|\mathcal{F}_q|_\textrm{C} + |\mathcal{F}_r|_\textrm{C}}{n^{\alpha_{qr}}}\right) + \varepsilon_r+\varepsilon_{qr} \leq O\left(\frac{|\mathcal{F}_q|_\textrm{C}}{\sqrt{n}}\right) + \varepsilon_q,
\end{equation}
which means when the number of training samples $n$ is the same, the upper bound of classification error based on the soft label is lower. When we want to achieve the same upper bound of classification error $R(f_{q}) - R(f)$ using these two techniques, learning from soft labels requires less data. This is the core reason why we use knowledge distillation and disagreement score $d_{\text{DS}}$ to select the coreset.

\section{Adaptive Coreset Selection for QAT} 

In Sec.~\ref{sec:importance}, we propose to use $d_{\text{EVS}}$ and $d_{\text{DS}}$ to select the coreset for QAT. While $d_{\text{DS}}$ could help select those samples that produce large performance gaps between quantized and full-precision models, $d_{\text{EVS}}$ targets more at the error of quantized prediction. These two metrics cover different characteristics of training data, and we need both to improve the diversity of our coreset.
For different stages of QAT, different metrics should be considered when selecting samples that fit the current training objective. Previous research~\citep{kim2019qkd} has shown that QAT should start with the hard label to help a better initialization for the quantized model and then use soft labels to guide it to better local minima. In light of this, we propose Adaptive Coreset Selection (ACS) for QAT to select the important samples considering current training epoch t, $d_{\text{EVS}}$, and $d_{\text{DS}}$ adaptively. 

For the given current training epoch $t$ and the total training epoch $E$, we propose a cosine annealing weights coefficient $\beta(t) = \cos (\frac{t}{2E} \pi)$ to consider two metrics simultaneously and balance between them. The final selection metric is a linear combination of $d_{\text{EVS}}$ and $d_{\text{DS}}$ as follows:
\begin{equation} \small
    d_{\text{ACS}}(t) = \beta(t)d_{\text{EVS}}(t) + (1-\beta(t))d_{\text{DS}}(t).
\end{equation}
As we have $\beta(0) = 1$ and $\beta(E) = 0$ in the early stage, the selection is mainly based on $d_{\text{EVS}}$. When the quantized model converges, we focus more on the $d_{\text{DS}}$ in later epochs. We perform coreset selection every $R$ epoch, where $R$ is determined before training. The pseudo-code for our ACS algorithm is shown in Alg.~\ref{alg:acs}. 
\begin{algorithm}[h] 
        \caption{Adaptive Coreset Selection for QAT}
       \label{alg:acs}
    \begin{algorithmic}
       \STATE {\bfseries Input:} Training dataset $T =\{(x_i,y_i)\}_{i=1}^n$, Real-value network with weights $\bf W^r$, Coreset data fraction per epoch $S$, Total training epochs $E$, Selection interval $R$, Initial coreset $T_{\text{ACS}}(t) = \emptyset$
       \STATE {\bfseries Output:} Quantized network with weights $\bf W^q$
       \STATE Initialize quantized weights $\bf W^q$ from $\bf W^r$ following Eq.~\ref{equ:quantizer}
       \FOR{$t \in [0, ..., E-1]$}
       \IF{$ t\%R == 0$} 
       \STATE $\beta(t) = \cos (\frac{t}{2E} \pi)$
       \FOR{$(x_i,y_i) \in T$}
       \STATE $d_{\text{EVS}}(x_i,t) = \| p(\bf W^q_t,x_i) - y_i \|_2$
       \STATE $d_{\text{DS}}(x_i,t) = \| p(\bf W^q_t,x_i) - p_\mathbf{T}(\bf W^r_t,x_i) \|_2$
       \STATE $d_{\text{ACS}}(x_i,t) = \beta(t)d_{\text{EVS}}(x_i,t) + (1-\beta(t))d_{\text{DS}}(x_i,t)$
       \ENDFOR
       \STATE Sort $d_{\text{ACS}}(x_i,t)$, select top $S\%$ samples to replace $T_{\text{ACS}}(t)$
       \ELSE
       \STATE $T_{\text{ACS}}(t) \gets T_{\text{ACS}}(t-1)$
       \ENDIF
       \STATE Train $\bf W^q$ on $T_{\text{ACS}}(t)$ following Eq.~\ref{eq:kd}
       \ENDFOR
    \end{algorithmic}
    \end{algorithm}

There are two major advantages of our quantization-aware ACS method. The first lies in the adaptation to the training phase when knowledge distillation is applied. As soft labels retain more information about the target than hard labels, we should encourage the quantized student model to learn sequentially on hard labels first and soft labels then. This implicit learning hierarchy is observed in QKD~\citep{kim2019qkd} and is named ``self-learning'' and ``tutoring''. With the proposed ACS fully aware of this hierarchy, the selected coreset helps stabilize the training and guarantee faster convergence. The second advantage is the diversity of training data. More samples could be covered in the coreset of different epochs, and the coverage of the original full dataset contributes to the convergence of a more robust model. Note that only when the \textbf{optimal data sequence} and \textbf{high training sample diversity} are achieved simultaneously, the performance of QAT will be significantly better. We demonstrate in the Appendix.~\ref{sec:coverage} that even when all data are covered but the order is random, the accuracy of our quantized model will be negatively influenced.


\section{Experiments}

\paragraph{Datasets and networks} The efficiency experiments are conducted on CIFAR-100~\citep{krizhevsky2009learning} and ImageNet-1K dataset~\citep{deng2009imagenet}. We evaluate MobileNetV2~\citep{howard2017mobilenets} on CIFAR-100 and evaluate ResNet-18~\citep{he2016deep} on the ImageNet-1K dataset. The \textit{width multiplier} is set to be 0.5 for MobileNetV2. We further provide experimental results of quantized RetinaNet~\cite{lin2017focal} on MS COCO object detection benchmark~\citep{lin2014microsoft}. The robustness experiments are conducted following the setting of \cite{paul2021el2n}. 
\paragraph{Baselines} We choose multiple coreset selection methods from different categories as our baseline. The selected methods include: Random Sampling, EL2N-Score~\citep{paul2021el2n}\footnote{Note that we only verify the EL2N-score instead of Gradient norm (GradN) score in \citet{paul2021el2n} based on the TMLR reproduction from \citet{kirsch2023does} observing that the inconsistency of GraNd at initialization will results in a sub-optimal performance even compared with random selection.}, Forgetting~\citep{toneva2018forgetting}, Glister~\citep{killamsetty2021glister}, kCenterGreedy~\citep{seneractive}, Contextual Diversity (CD)~\citep{agarwal2020contextual}, Moderate Coreset~\citep{xia2023moderate}. For methods involving early training, we set training epochs as 5. Note that comparing adaptive-based and non-adaptive-based methods is fair and the common practice of previous research~\citep{pooladzandi2022adaptive, killamsetty2021retrieve}. 
\paragraph{Training and selection details} For MobileNetV2, we train the network for 200 epochs using a learning rate of 0.01, weight decay of 5e-4, batch size of 512, $R=20$, and SGD optimizer. For ResNet-18 on ImageNet-1K, we train the network for 120 epochs using a learning rate of 1.25e-3, no weight decay, batch size of 512, $R=10$, and Adam optimizer. We use the quantization method and full-precision model following LSQ+~\citep{bhalgat2020lsq+}. All the experiments were carried out on 2 NVIDIA RTX 3090 GPUs. As we notice that the results on the ImageNet-1K dataset will not vary significantly across different runs, experiments of each setting are only performed once. For CIFAR-100 experiments, each experiment of different settings is repeated 5 times. For a fair comparison, we use knowledge distillation with the corresponding full-precision model as the teacher in \textbf{all} experiments, regardless of the method and dataset fraction. 
\begin{table*}[ht]
  \caption{Comparison of applying different coreset selection methods to 2/32-bit quantized MobileNetV2 on CIFAR-100 dataset with various subset fractions. When full data are selected ($\mathcal{S}=100\%$), the mean accuracy and standard deviation is 68.1$\pm$0.9\%. The first and second best accuracy is shown in \textbf{bold} and \underline{underlined}, and the performance improvement over the previous SOTA is marked in \textcolor{blue}{blue}.}
  \label{tab:mbnetv2}
  \centering
  \resizebox{\textwidth}{!}{
  \begin{tabular}{l|ccccc}
    \toprule
    Method/Fraction   & $\mathcal{S}=10\%$ & $\mathcal{S}=20\%$ & $\mathcal{S}=30\%$ & $\mathcal{S}=40\%$ &$\mathcal{S}=50\%$ \\
    \midrule
    Random                                 &62.3$\pm$0.9 &\underline{64.1$\pm$0.7} &65.2$\pm$0.3 &\underline{65.6$\pm$0.6} &66.2$\pm$0.9  \\
    EL2N Score~\citep{paul2021el2n}        &\underline{63.0$\pm$0.6} &64.0$\pm$1.6 &\underline{65.6$\pm$0.8} &64.6$\pm$0.8 &65.3$\pm$0.4    \\
    Forgetting~\citep{toneva2018forgetting} &60.7$\pm$0.3 &63.1$\pm$0.6 &65.0$\pm$0.8 &65.3$\pm$0.8 &66.0$\pm$1.2  \\
    Glister~\citep{killamsetty2021glister}  &56.5$\pm$1.2 &60.4$\pm$1.0 &61.3$\pm$0.6 &63.0$\pm$1.0 &64.8$\pm$0.9  \\
    kCenterGreedy~\citep{seneractive}       &60.2$\pm$0.8 &62.7$\pm$0.5 &63.8$\pm$0.7 &64.8$\pm$0.6 &\underline{66.3$\pm$1.1}  \\
    CD~\citep{agarwal2020contextual} &60.3$\pm$0.8&62.6$\pm$0.9&64.0$\pm$1.0&64.8$\pm$0.4&65.3$\pm$0.4\\
    Moderate~\citep{xia2023moderate}        &57.9$\pm$0.3 &60.4$\pm$0.7 &62.8$\pm$0.6 &63.6$\pm$0.8 &64.5$\pm$1.6  \\
    \midrule
    Ours           &\textbf{63.7$\pm$0.8} (\textcolor{blue}{$\uparrow$0.7})& \textbf{65.9$\pm$0.7} (\textcolor{blue}{$\uparrow$1.8})& \textbf{66.4$\pm$0.8} (\textcolor{blue}{$\uparrow$0.8})& \textbf{66.9$\pm$0.5} (\textcolor{blue}{$\uparrow$1.3})& \textbf{67.2$\pm$0.5}(\textcolor{blue}{$\uparrow$0.9}) \\
    \bottomrule
  \end{tabular}}
\end{table*}

\begin{table*}[ht]
  \caption{Comparison of different methods of 4/4-bit quantized ResNet-18 on ImageNet-1K. When full data are selected ($\mathcal{S}=100\%$), the accuracy is 72.46\%. The first and second best accuracy is shown in \textbf{bold} and \underline{underlined}, and the performance improvement over the previous SOTA is marked in \textcolor{blue}{blue}.}
  \label{tab:resnet18}
  \centering
  \resizebox{\textwidth}{!}{
  \begin{tabular}{l|cccccc}
    \toprule
    Method/Fraction (\%)   & $\mathcal{S}=10\%$ &$\mathcal{S}=30\%$ & $\mathcal{S}=50\%$ & $\mathcal{S}=60\%$ & $\mathcal{S}=70\%$ & $\mathcal{S}=80\%$  \\
    \midrule
    Random         &\underline{64.15} &68.53 &70.49 &70.94 & 71.06 &71.96  \\
    EL2N Score~\citep{paul2021el2n}          &61.71 & 67.31 & 70.14&70.89&71.54&71.88\\
    Forgetting~\citep{toneva2018forgetting}  &63.09 & 67.77 & 70.14&71.00 &71.36&71.82\\
    Glister~\citep{killamsetty2021glister}   &63.25 & \underline{68.94} & \underline{70.92} & \underline{71.39} & \underline{71.93} & \underline{72.22} \\
    kCenterGreedy~\citep{seneractive}        &62.98 & 68.56 & 70.35&71.13&71.59&71.96 \\
    CD~\citep{agarwal2020contextual}         &63.22 & 68.74 & 70.90&71.27&71.78&72.11\\
    Moderate~\citep{xia2023moderate}         &62.39 & 68.06 & 70.43&70.62&71.56&71.99\\
    \midrule
    Ours    & \textbf{68.39} (\textcolor{blue}{$\uparrow$4.24}) & \textbf{71.09} (\textcolor{blue}{$\uparrow$2.15}) & \textbf{71.59} (\textcolor{blue}{$\uparrow$0.67})& \textbf{72.00} (\textcolor{blue}{$\uparrow$0.61})& \textbf{72.19} (\textcolor{blue}{$\uparrow$0.26})& \textbf{72.31} (\textcolor{blue}{$\uparrow$0.09})\\
    \bottomrule
  \end{tabular}}
\end{table*}

\subsection{Benchmarking Previous Coreset Method}
The comparison of the QAT Top-1 accuracy of MobileNetV2 on CIFAR-100 and ResNet-18 on ImageNet-1K is shown in Tab.~\ref{tab:mbnetv2} and Tab.~\ref{tab:resnet18}. We note that most previous methods cannot exceed random selection in our QAT setting, and the few surpass the baseline only on specific data fractions. This trend is also discussed for full-precision training by \citet{kirsch2023does}. For example, kCenterGreedy~\citep{seneractive} shows satisfying performance when the subset size is large (50\% for CIFAR-10, 70\%/80\% for ImageNet-1K) but fails to demonstrate effectiveness on small coreset size. Our method outperforms state-of-the-art methods on all subset fractions $\mathcal{S}$ by a great margin. Specifically, the ResNet-18 accuracy of 10\% subset on ImageNet-1K using our method is 68.39\%, achieving an absolute gain of 4.24\% compared to the baseline method. 

\paragraph{Efficiency analysis}
The detailed QAT training time of ResNet-18 on ImageNet-1K coreset and 2 NVIDIA RTX 3090 GPUs with different methods is listed in Tab.~\ref{tab:efficiency}. When full data are selected ($\mathcal{S}=100\%$), the training time is 62.3h. The only method with comparable efficiency to ours is the Moderate Coreset~\citep{xia2023moderate}, which only needs to perform forwarding on samples and can eliminate optimization or greedy searching. The results prove that our method can effectively reduce training time without incurring computation overhead by the selection algorithm.
\begin{table}[h]
  \caption{Training time (hours) of 4/4-bit quantized ResNet-18 on ImageNet-1K with various subset fraction.}
  \label{tab:efficiency}
  \centering
  \resizebox{0.6\textwidth}{!}{ 
  \begin{tabular}{l|cccccc}
    \toprule
    Method/Fraction (\%)   & 10\% &30\% &50\% &60\% &70\% &80\%  \\
    \midrule
    EL2N Score          &12.7 & 23.8 & 39.1 & 44.1 & 49.9 & 56.0\\
    Forgetting  &12.7 & 23.9 & 39.5 & 44.7 & 50.2 & 56.6\\
    Glister   &13.8 & 23.0 & 39.9 & 46.6 & 58.0 & 66.8\\
    kCenterGreedy        &13.0 & 25.2 & 38.4 & 43.8 & 50.5 & 56.9 \\
    CD         &11.9 & 24.5 & 37.1 & 42.5 & 48.9 & 55.0 \\
    Moderate         &11.3 & 22.3 & 36.2 & 42.5 & 48.2  & 54.1\\
    \midrule
    Ours (R=10)   & \textbf{11.2} & \textbf{20.7} & \textbf{36.1} & \textbf{42.2} & \textbf{48.0} & \textbf{53.7} \\
    \bottomrule
  \end{tabular}}
\end{table}
\paragraph{Effect of adaptive epochs}
The choice of selection interval $R$ is vital in our algorithm, as too large $R$ will fail to help adaption and improve data diversity, and too small $R$ will introduce too much computation overhead and undermine the efficiency. We apply grid search on $R$ and empirically prove that $R=10$ is optimal for our ImageNet-1K training. The accuracy and training time results are shown in Tab.~\ref{tab:adaptive_epoch}. We can observe from the results that $R=10$ achieves a similar performance as $R=5$ with a shorter training time. As no back-propagation of the gradient is involved during the computation of $d_{\text{EVS}}$ and $d_{\text{DS}}$, the computation overhead is acceptable under most settings with $R>5$ compared with previous methods involving training. 
\begin{table}[h]
\begin{minipage}[b]{0.58\textwidth}
  \caption{Analysis on adaptive epochs $R$.}
  \label{tab:adaptive_epoch}
  \centering
  \resizebox{\textwidth}{!}{
  \begin{tabular}{l|cc|cc|cc|cc}
    \toprule
    \multirow{2}*{Fraction}   & \multicolumn{2}{c}{$\mathcal{S}$=10\%} &\multicolumn{2}{c}{$\mathcal{S}$=30\%} & \multicolumn{2}{c}{$\mathcal{S}$=50\%}  & \multicolumn{2}{c}{$\mathcal{S}$=70\%} \\
    & Acc. & Time & Acc. & Time & Acc. & Time & Acc. & Time \\
    \midrule
    $R=5$ & 68.8 & 12.7h&71.05 & 21.5h& 71.47 & 37.0h& 72.15 & 49.1h\\
    $R=10$ & 68.39 & 11.3h& 71.09 & 20.7h& 71.59 & 36.1h& 72.19 &48.0h\\
    $R=20$ & 67.58 & 11.0h& 70.35 & 20.2h& 71.42 & 35.5h& 71.97 &47.6h\\
    $R=40$ & 66.10 & 10.7h& 69.86 & 19.8h & 71.25 & 34.9h& 72.00 & 47.2h\\
    $R=60$ & 64.96 & 10.5h& 69.27 & 19.5h & 71.05 & 34.4h& 71.93 & 46.9h\\
    $R>120$ & 62.82 & 10.3h& 67.62 & 19.2h & 69.98 & 33.9h& 71.37 & 46.6h\\
    \bottomrule
  \end{tabular}}
  \end{minipage}
  \begin{minipage}[b]{0.38\textwidth}
  \caption{Analysis on strategy $\beta(t)$.}
  \label{tab:beta}
  \centering
  \resizebox{\textwidth}{!}{
  \begin{tabular}{lcccc}
    \toprule
    Fraction (\%)  &  10\%&  30\%&  50\%&  70\%\\

    \midrule
    fixed     &  67.95 & 70.83 & 71.21 & 72.01\\
    linear    &  68.35 & 71.07 & 71.39 & 72.11\\
    sqrt      &  68.15 & 71.03 & 71.44 & 72.15\\
    quadratic &  68.11 & 70.95 & 71.35 & 72.10\\
    \midrule
    $d_{\text{EVS}}$ only & 68.06 & 70.63 & 71.41 & 71.96\\
    $d_{\text{DS}}$ only & 67.07 & 70.24 & 71.43 & 72.08 \\
    \midrule
    cosine & \textbf{68.39} & \textbf{71.09} & \textbf{71.59} &\textbf{72.19} \\
    \bottomrule
  \end{tabular}}
  \end{minipage}
\end{table}

\begin{figure*}[t]
  \centering
  \begin{subfigure}{0.26\linewidth}
    \includegraphics[width=\textwidth]{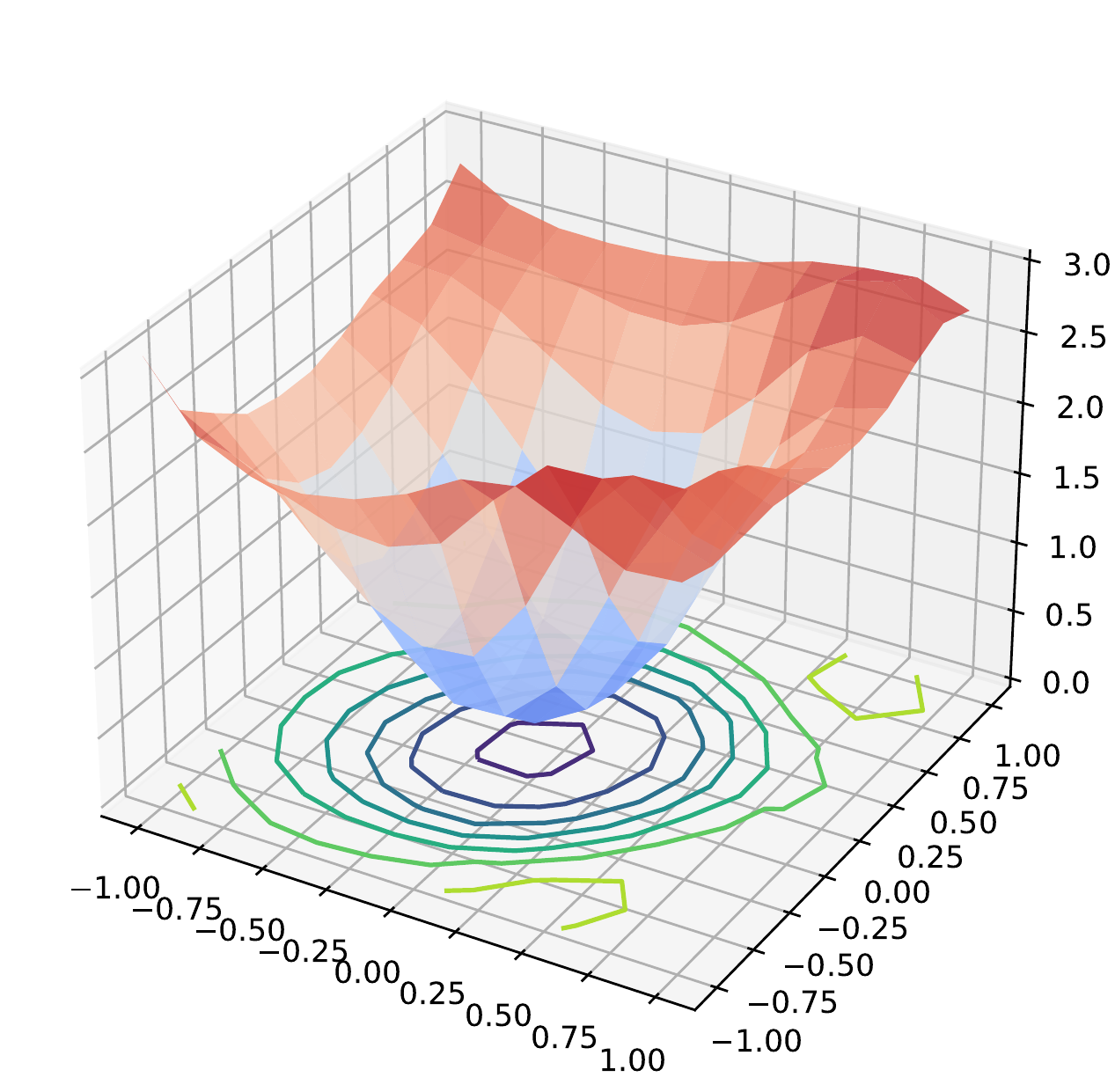}
    \caption{Full Dataset}
    \label{fig:mbnetv2-full}
  \end{subfigure}
  \begin{subfigure}{0.25\linewidth}
    \includegraphics[width=\textwidth]{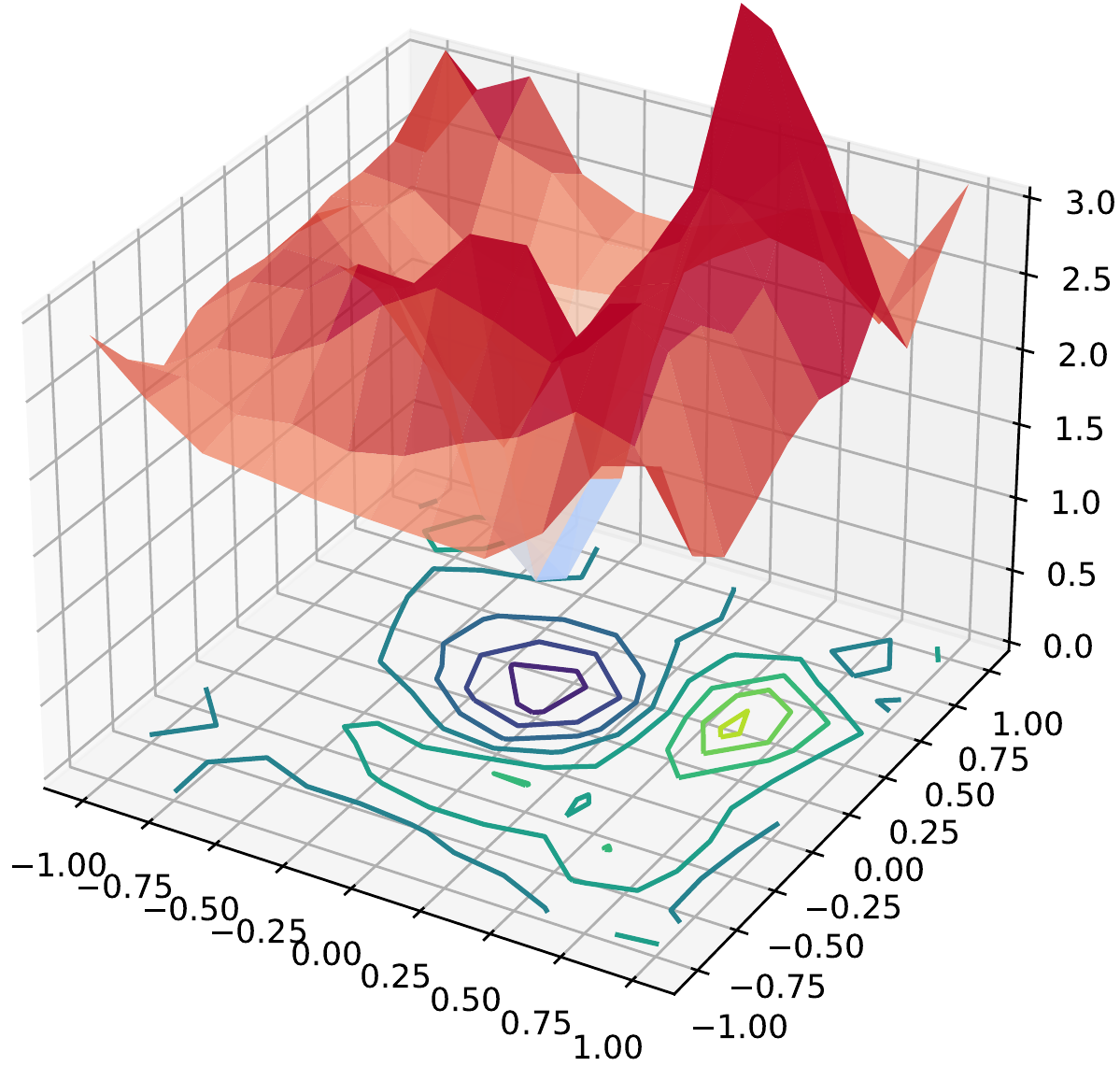}
    \caption{Random $\mathcal{S}=10\%$}
    \label{fig:mbnetv2-random}
  \end{subfigure}
  \begin{subfigure}{0.31\linewidth}
    \includegraphics[width=\textwidth]{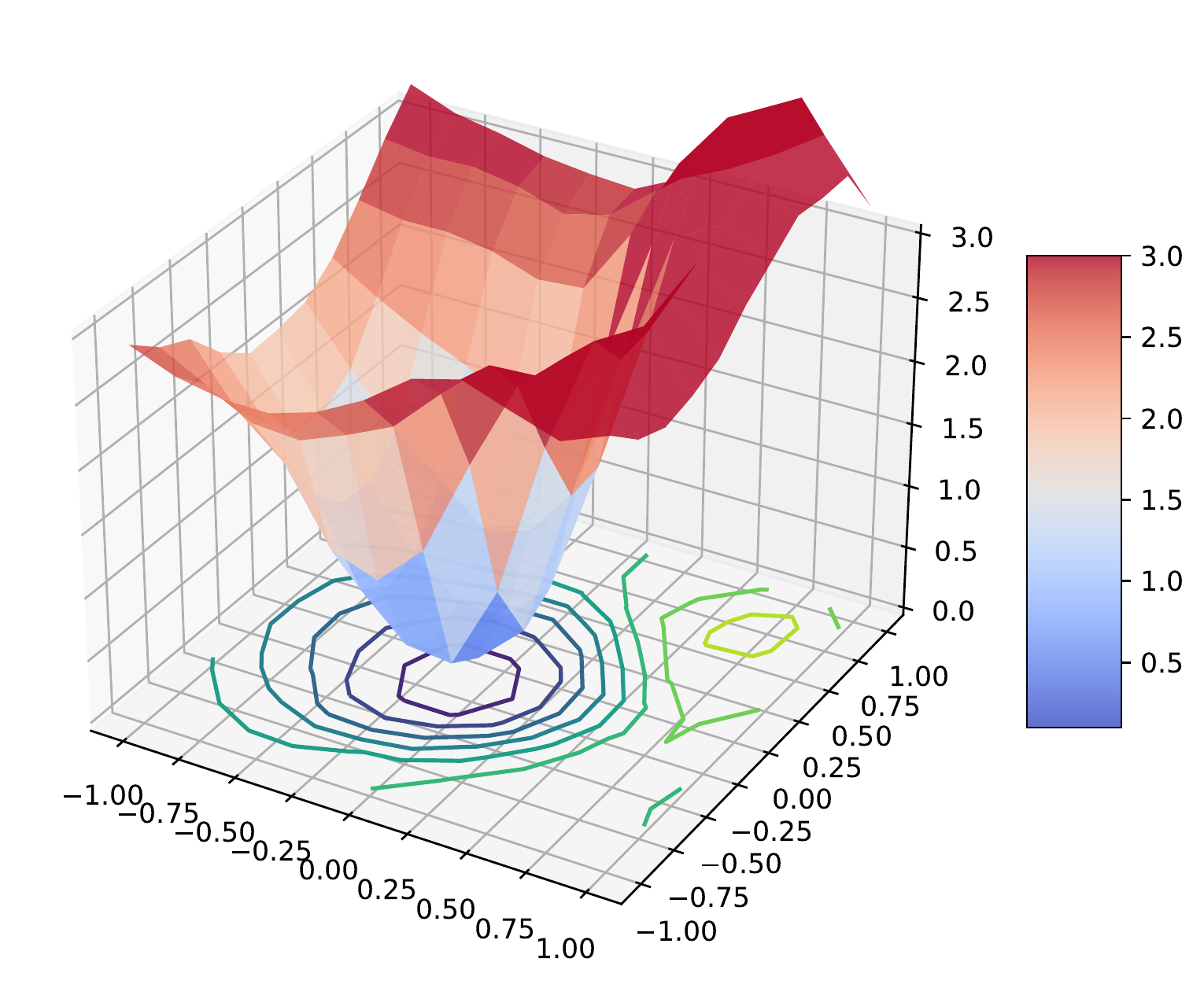}
    \caption{Ours $\mathcal{S}=10\%$}
    \label{fig:mbnetv2-ours}
  \end{subfigure}
  \caption{Visualization of the loss landscape~\citep{li2018visualizing} of 2-bit quantized MobileNetV2 trained on the CIFAR-100 (a) full dataset, (b) 10\% random subset, and (c) 10\% ACS coreset.}
  \label{fig:losslandscape}
\end{figure*}

\paragraph{Ablation study and effect of annealing strategy} 

As we propose two metrics for coreset selection: $d_{\text{EVS}}$ and $d_{\text{DS}}$, it is important to analyze how we should balance between them and the contribution of both metrics. We try the following different annealing strategies and settings: (1) fixed ($\beta(t) = 0.5$); (2) linear ($\beta(t) = 1-\frac{t}{E}$); (3) square root ($\beta(t) = 1-\sqrt{\frac{t}{E}}$); (4) quadratic ($\beta(t) = 1-(\frac{t}{E})^2$); (5) cosine ($\beta(t) = \cos (\frac{t}{2E} \pi)$); (6) $d_{\text{EVS}}$ only ($\beta(t) = 1$); (7) $d_{\text{DS}}$ only ($\beta(t) = 0$); The results of coreset selection of 4-bit quantized ResNet-18 on ImageNet-1K are listed in Tab.~\ref{tab:beta}. The performance gap with different annealing strategies is close as they all follow the trends to use $d_{\text{EVS}}$ in the early epochs of training and $d_{\text{DS}}$ in later epochs. Among all these annealing strategies, cosine annealing is slightly better. When only one metric is used for the selection, the performance will drop. 
We also notice that $d_{\text{EVS}}$ and $d_{\text{DS}}$ are complementary as $d_{\text{EVS}}$ works well with small data fraction and $d_{\text{DS}}$ has better performance with large data fraction.

\subsection{Training with Label Noise}
We would like to highlight another application of our method: identifying noisy labels in training data and improving the robustness of the QAT. The nature of advantage of our method is to select important training examples for QAT and remove those low-quality redundant examples. This is especially useful when there is label noise in the training set (some examples are not correctly labeled). This application is proposed in previous coreset research~\citet{mirzasoleiman2020coresets}, but not verified on QAT. We follow the setting of~\cite{mirzasoleiman2020coresets} to experiment on the QAT of ResNet-18 on CIFAR-10 with 10\% randomized labels (labels of 10\% of training samples are randomly re-generated). The ResNet-18 is quantized to 2/32 for weights and activations. When the full dataset is selected, the accuracy is 89.91$\pm$0.51\%. The comparison of the QAT Top-1 accuracy (\%) of 2/32-bit quantized ResNet-18 on CIFAR-10 with 10\% randomized labels is shown as follows in Tab.~\ref{tab:noise}. Note that we select E2LN Score and GraNd Score for comparison as they are the few coreset selection solutions among the methods we investigate in Tab.~\ref{tab:resnet18} claimed with the capacity to remove noise in the training set and improve training robustness.

\begin{table}[h]
  \caption{Top-1 Accuracy of 2/32-bit quantized ResNet-18 on CIFAR-10 with 10\% random label noise}
  \label{tab:noise}
  \centering
  \resizebox{0.8\textwidth}{!}{ 
  \begin{tabular}{lcccccc}
    \toprule
    Method/Fraction (\%)   & 10\% &20\% &30\% &40\% &50\%\\
    \midrule
    Random                 &83.14$\pm$0.51 &84.77$\pm$0.80 &84.82$\pm$0.57 &85.09$\pm$0.15 & 85.33$\pm$0.60   \\
    EL2N Score             &85.80$\pm$0.51 & 86.02$\pm$0.55 & 87.18$\pm$0.35 &87.51$\pm$0.29 & 88.01$\pm$0.45\\
    GraNd Score             &85.71$\pm$0.30 & 85.96$\pm$0.15 & 87.10$\pm$0.55 &87.44$\pm$0.20 & 87.94$\pm$0.29\\
    \midrule
    Ours    &\textbf{88.31$\pm$0.27}&\textbf{89.53$\pm$0.62}&\textbf{89.95$\pm$0.39}&\textbf{90.21$\pm$0.39}&\textbf{90.23$\pm$0.18} \\
    \bottomrule
  \end{tabular}}
\end{table}

As can be seen from the results in Tab.~\ref{tab:noise}, our method outperforms all other selection baselines and even performs better than full-data training when coreset size $\mathcal{S}\geq30\%$. To quantitatively demonstrate the noisy sample pruning effectiveness, we report the average recall for this noisy-sample detection problem. Noticeably, while the other pruned samples are in the correct label, they are still mostly redundant and can be pruned without significant performance degradation in QAT. From the results of recall in Tab~\ref{tab:noise_recall}, our method states most of the noisy samples. These results show that our method can successfully prune those samples with incorrect labels. In this case, we actually achieve a \textbf{``lossless acceleration''} as both the accuracy and efficiency are better than full-data training. 

\begin{table}[h]
  \caption{The average recall of noisy sample identification on CIFAR-10 with 10\% random label noise.}
  \label{tab:noise_recall}
  \centering
  \resizebox{0.55\textwidth}{!}{ 
  \begin{tabular}{lcccccc}
    \toprule
    Method/Fraction (\%)   & 10\% &20\% &30\% &40\% &50\%\\
    \midrule
    EL2N Score             & 90.7\% & 80.5\% & 71.0\% & 60.2\% & 47.4\% \\
    GraNd Score            & 92.3\% & 81.7\% & 65.4\% & 41.0\% & 26.5\% \\
    \midrule
    Ours    & 97.9\% & 90.2\% & 84.1\% & 78.9\% & 73.5\% \\
    \bottomrule
  \end{tabular}}
\end{table}

\subsection{Object Detection Tasks} 

We further conduct experiments of 4/4-bit QAT of RetinaNet~\citep{lin2017focal} with ResNet-18 backbone on the MS COCO object detection dataset~\citep{lin2014microsoft}. We use the same selection metric for the RetinaNet training and use the output for the classification head as the probability vector $p$. When there are multiple objects in one image, we use the mean selection metric $d_{\text{ACS}}$ of the probability output of all objects. The QAT method used is FQN~\citep{li2019fully} and we perform training on \texttt{coco-2017-train} for 100 epochs. We investigate two coreset fractions: 10\% and 50\% and use $R$=10. The results of mAP are listed in Tab.~\ref{tab:detection}. Our method outperforms the random selection baseline and state-of-the-art selection method Moderate~\citep{xia2023moderate} significantly, which proves that our method is effective on object detection tasks.

\begin{table}[h]
  \caption{Comparision of performance with different methods of RetinaNet on COCO benchmarks.}
  \label{tab:detection}
  \centering
  \resizebox{0.6\textwidth}{!}{
  \begin{tabular}{lccccccc}
    \toprule
    Method & Fraction  & AP & $\text{AP}^{0.5}$ & $\text{AP}^{0.75}$ & $\text{AP}^{\text{S}}$ & $\text{AP}^{\text{M}}$ & $\text{AP}^{\text{L}}$\\
    \midrule
    Full & 100\% & 28.6 & 46.9 & 29.9 & 14.9 & 31.2 & 38.7 \\
    \midrule
    Random   & 10\% & 21.4&39.8&22.4&7.5&24.3&27.4 \\
    EL2N Score& 10\% & 20.7&37.0&19.9&8.4&25.0&28.1 \\
    Moderate & 10\% & 22.0&37.8&20.4&8.4&25.0&28.1 \\
    Ours     & 10\% & 24.4&40.9&25.1&9.9&27.1&31.5  \\
    \midrule
    Random   & 50\% & 25.4&42.5&26.7&10.7&27.7&32.1 \\
    EL2N Score& 50\% & 25.9&42.8&27.0&11.4&28.6&33.8 \\
    Moderate & 50\% & 25.0&42.7&25.9&11.2&28.5&33.0  \\
    Ours     & 50\% & 26.7&44.0&27.8&12.1&30.0&35.1
  \\
    \bottomrule
  \end{tabular}}
\end{table}

\subsection{Visualization and Analysis} 
\begin{figure}[H]
  \centering
  \begin{subfigure}{0.41\linewidth}
    \includegraphics[width=\textwidth]{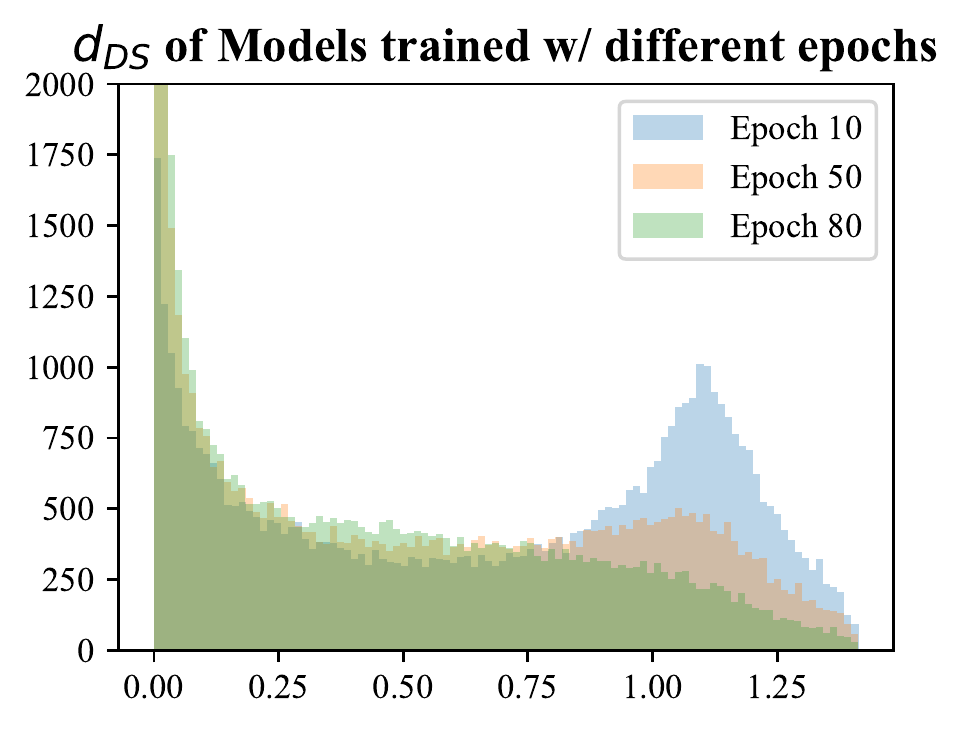}
    \caption{$d_{\text{DS}}$ distribution}
    \label{fig:score1}
  \end{subfigure}
  \begin{subfigure}{0.4\linewidth}
    \includegraphics[width=\textwidth]{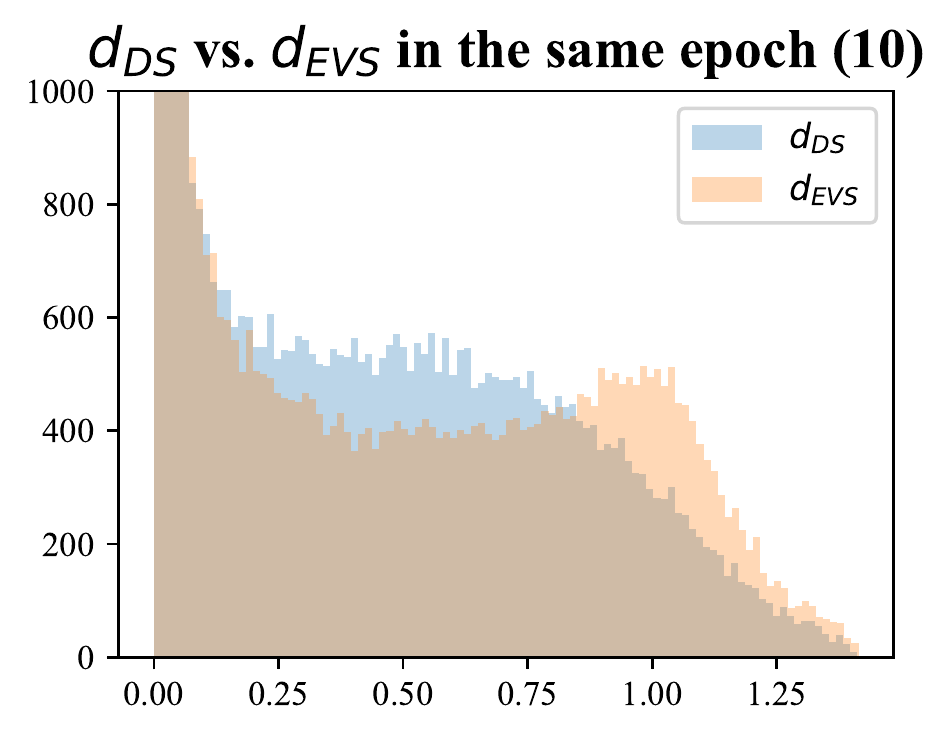}
    \caption{$d_{\text{DS}}$ vs. $d_{\text{EVS}}$}
    \label{fig:score2}
  \end{subfigure}
  \caption{(a) Distribution of disagreement scores $d_{\text{DS}}$ on MobileNetV2 for different epochs. (b) Distribution of disagreement scores $d_{\text{DS}}$ and error vector score $d_{\text{EVS}}$ on MobileNetV2 in the same epoch.}
  \label{fig:top1acc}
\end{figure}

We visualize the loss landscape~\citep{li2018visualizing} of MobileNetV2 training on the full CIFAR-100 dataset, 10\% random subset of CIFAR-100, and 10\% coreset of CIFAR-100 based on our methods shown in Fig.~\ref{fig:losslandscape}. We can see from the results that QAT on coreset with our method has a more centralized and smoother loss compared to the baseline methods, which reflects that our method helps improve the training stability of QAT. We also visualize the distribution of disagreement scores $d_{\text{DS}}$ and error vector score $d_{\text{EVS}}$ in Fig.~\ref{fig:score1} and Fig.~\ref{fig:score2}. The setting is the same as the MobileNetV2 experiment listed in Tab.~\ref{tab:mbnetv2}. We can see from the results that the mean of $d_{\text{DS}}$ shifts to zero during the QAT, which proves that $d_{\text{DS}}$ is a useful metric to quantify the importance of each sample. The distribution discrepancy between $d_{\text{EVS}}$ and $d_{\text{DS}}$ proves the necessity of considering both metrics to select diverse data into our coreset.

\section{Conclusion}
This is the first work focusing on the data efficiency and robustness of quantization-aware training (QAT).
By removing samples from the training batches and analyzing the loss gradient, we theoretically prove that the importance of each sample varies significantly for QAT. The error-vector score and disagreement score are proposed to quantify this importance. Considering the training characteristics of QAT, we propose a fully quantization-aware Adaptive Coreset Selection (ACS) method to better adapt to different training phases and improve training data diversity. Extensive experiments on various datasets, networks, and quantization settings further demonstrate the effectiveness of our method. We also verified the proposed ACS on the noisy label setting and proved that our method can improve the robustness of QAT.

\section*{Acknowledgement}
This research is supported by HKSAR RGC General Research Fund (GRF) \#16208823.

\bibliography{main}

\begin{thebibliography}{92}
\providecommand{\natexlab}[1]{#1}
\providecommand{\url}[1]{\texttt{#1}}
\expandafter\ifx\csname urlstyle\endcsname\relax
  \providecommand{\doi}[1]{doi: #1}\else
  \providecommand{\doi}{doi: \begingroup \urlstyle{rm}\Url}\fi

\bibitem[Agarwal et~al.(2020)Agarwal, Arora, Anand, and Arora]{agarwal2020contextual}
Sharat Agarwal, Himanshu Arora, Saket Anand, and Chetan Arora.
\newblock Contextual diversity for active learning.
\newblock In \emph{Computer Vision--ECCV 2020: 16th European Conference, Glasgow, UK, August 23--28, 2020, Proceedings, Part XVI}, pp.\  137--153, 2020.

\bibitem[Alabdulmohsin et~al.(2022)Alabdulmohsin, Neyshabur, and Zhai]{alabdulmohsin2022revisiting}
Ibrahim~M Alabdulmohsin, Behnam Neyshabur, and Xiaohua Zhai.
\newblock Revisiting neural scaling laws in language and vision.
\newblock \emph{Advances in Neural Information Processing Systems}, 35:\penalty0 22300--22312, 2022.

\bibitem[Bengio et~al.(2013)Bengio, L{\'e}onard, and Courville]{bengio2013estimating}
Yoshua Bengio, Nicholas L{\'e}onard, and Aaron Courville.
\newblock Estimating or propagating gradients through stochastic neurons for conditional computation.
\newblock \emph{arXiv preprint arXiv:1308.3432}, 2013.

\bibitem[Bhalgat et~al.(2020)Bhalgat, Lee, Nagel, Blankevoort, and Kwak]{bhalgat2020lsq+}
Yash Bhalgat, Jinwon Lee, Markus Nagel, Tijmen Blankevoort, and Nojun Kwak.
\newblock Lsq+: Improving low-bit quantization through learnable offsets and better initialization.
\newblock In \emph{Proceedings of the IEEE/CVF Conference on Computer Vision and Pattern Recognition Workshops}, pp.\  696--697, 2020.

\bibitem[Borsos et~al.(2020)Borsos, Mutny, and Krause]{borsos2020coresets}
Zal{\'a}n Borsos, Mojmir Mutny, and Andreas Krause.
\newblock Coresets via bilevel optimization for continual learning and streaming.
\newblock \emph{Advances in Neural Information Processing Systems}, 33:\penalty0 14879--14890, 2020.

\bibitem[Brown et~al.(2020)Brown, Mann, Ryder, Subbiah, Kaplan, Dhariwal, Neelakantan, Shyam, Sastry, Askell, et~al.]{brown2020language}
Tom Brown, Benjamin Mann, Nick Ryder, Melanie Subbiah, Jared~D Kaplan, Prafulla Dhariwal, Arvind Neelakantan, Pranav Shyam, Girish Sastry, Amanda Askell, et~al.
\newblock Language models are few-shot learners.
\newblock \emph{Advances in neural information processing systems}, 33:\penalty0 1877--1901, 2020.

\bibitem[Choi et~al.(2018)Choi, Wang, Venkataramani, Chuang, Srinivasan, and Gopalakrishnan]{choi2018pact}
Jungwook Choi, Zhuo Wang, Swagath Venkataramani, Pierce I-Jen Chuang, Vijayalakshmi Srinivasan, and Kailash Gopalakrishnan.
\newblock Pact: Parameterized clipping activation for quantized neural networks.
\newblock \emph{arXiv preprint arXiv:1805.06085}, 2018.

\bibitem[Conneau \& Lample(2019)Conneau and Lample]{conneau2019cross}
Alexis Conneau and Guillaume Lample.
\newblock Cross-lingual language model pretraining.
\newblock \emph{Advances in neural information processing systems}, 32, 2019.

\bibitem[Deng et~al.(2009)Deng, Dong, Socher, Li, Li, and Fei-Fei]{deng2009imagenet}
Jia Deng, Wei Dong, Richard Socher, Li-Jia Li, Kai Li, and Li~Fei-Fei.
\newblock Imagenet: A large-scale hierarchical image database.
\newblock In \emph{2009 IEEE conference on computer vision and pattern recognition}, pp.\  248--255. Ieee, 2009.

\bibitem[Dosovitskiy et~al.(2020)Dosovitskiy, Beyer, Kolesnikov, Weissenborn, Zhai, Unterthiner, Dehghani, Minderer, Heigold, Gelly, et~al.]{dosovitskiy2020image}
Alexey Dosovitskiy, Lucas Beyer, Alexander Kolesnikov, Dirk Weissenborn, Xiaohua Zhai, Thomas Unterthiner, Mostafa Dehghani, Matthias Minderer, Georg Heigold, Sylvain Gelly, et~al.
\newblock An image is worth 16x16 words: Transformers for image recognition at scale.
\newblock \emph{arXiv preprint arXiv:2010.11929}, 2020.

\bibitem[Ducoffe \& Precioso(2018)Ducoffe and Precioso]{ducoffe2018adversarial}
Melanie Ducoffe and Frederic Precioso.
\newblock Adversarial active learning for deep networks: a margin based approach.
\newblock \emph{arXiv preprint arXiv:1802.09841}, 2018.

\bibitem[Esser et~al.(2020)Esser, McKinstry, Bablani, Appuswamy, and Modha]{esser2020learned}
Steven~K Esser, Jeffrey~L McKinstry, Deepika Bablani, Rathinakumar Appuswamy, and Dharmendra~S Modha.
\newblock Learned step size quantization.
\newblock In \emph{International Conference on Learning Representations}, 2020.

\bibitem[Fang et~al.(2020)Fang, Shafiee, Abdel-Aziz, Thorsley, Georgiadis, and Hassoun]{fang2020post}
Jun Fang, Ali Shafiee, Hamzah Abdel-Aziz, David Thorsley, Georgios Georgiadis, and Joseph~H Hassoun.
\newblock Post-training piecewise linear quantization for deep neural networks.
\newblock In \emph{European Conference on Computer Vision}, pp.\  69--86. Springer, 2020.

\bibitem[Fort \& Ganguli(2019)Fort and Ganguli]{fort2019emergent}
Stanislav Fort and Surya Ganguli.
\newblock Emergent properties of the local geometry of neural loss landscapes.
\newblock \emph{arXiv preprint arXiv:1910.05929}, 2019.

\bibitem[Fort et~al.(2020)Fort, Dziugaite, Paul, Kharaghani, Roy, and Ganguli]{fort2020deep}
Stanislav Fort, Gintare~Karolina Dziugaite, Mansheej Paul, Sepideh Kharaghani, Daniel~M Roy, and Surya Ganguli.
\newblock Deep learning versus kernel learning: an empirical study of loss landscape geometry and the time evolution of the neural tangent kernel.
\newblock \emph{Advances in Neural Information Processing Systems}, 33:\penalty0 5850--5861, 2020.

\bibitem[Fujishige(2005)]{fujishige2005submodular}
Satoru Fujishige.
\newblock \emph{Submodular functions and optimization}.
\newblock Elsevier, 2005.

\bibitem[He et~al.(2016)He, Zhang, Ren, and Sun]{he2016deep}
Kaiming He, Xiangyu Zhang, Shaoqing Ren, and Jian Sun.
\newblock Deep residual learning for image recognition.
\newblock In \emph{Proceedings of the IEEE conference on computer vision and pattern recognition}, pp.\  770--778, 2016.

\bibitem[Hinton et~al.(2015)Hinton, Vinyals, and Dean]{hinton2015distilling}
Geoffrey Hinton, Oriol Vinyals, and Jeff Dean.
\newblock Distilling the knowledge in a neural network.
\newblock \emph{arXiv preprint arXiv:1503.02531}, 2015.

\bibitem[Howard et~al.(2017)Howard, Zhu, Chen, Kalenichenko, Wang, Weyand, Andreetto, and Adam]{howard2017mobilenets}
Andrew~G Howard, Menglong Zhu, Bo~Chen, Dmitry Kalenichenko, Weijun Wang, Tobias Weyand, Marco Andreetto, and Hartwig Adam.
\newblock Mobilenets: Efficient convolutional neural networks for mobile vision applications.
\newblock \emph{arXiv preprint arXiv:1704.04861}, 2017.

\bibitem[Huang et~al.(2022)Huang, Shen, Li, Liu, Xianghong, Wicaksana, Xing, and Cheng]{huang2022sdq}
Xijie Huang, Zhiqiang Shen, Shichao Li, Zechun Liu, Hu~Xianghong, Jeffry Wicaksana, Eric Xing, and Kwang-Ting Cheng.
\newblock Sdq: Stochastic differentiable quantization with mixed precision.
\newblock In \emph{International Conference on Machine Learning}, pp.\  9295--9309. PMLR, 2022.

\bibitem[Huang et~al.(2023)Huang, Zhang, and Shan]{huang2023twin}
Zhizhong Huang, Junping Zhang, and Hongming Shan.
\newblock Twin contrastive learning with noisy labels.
\newblock In \emph{Proceedings of the IEEE/CVF Conference on Computer Vision and Pattern Recognition}, pp.\  11661--11670, 2023.

\bibitem[Iyer et~al.(2020)Iyer, Khargoankar, Bilmes, and Asanani]{iyer2020submodular}
Rishabh Iyer, Ninad Khargoankar, Jeff Bilmes, and Himanshu Asanani.
\newblock Submodular combinatorial information measures with applications in machine learning.
\newblock \emph{arXiv preprint arXiv:2006.15412}, 2020.

\bibitem[Jouppi et~al.(2017)Jouppi, Young, Patil, Patterson, Agrawal, Bajwa, Bates, Bhatia, Boden, Borchers, et~al.]{jouppi2017datacenter}
Norman~P Jouppi, Cliff Young, Nishant Patil, David Patterson, Gaurav Agrawal, Raminder Bajwa, Sarah Bates, Suresh Bhatia, Nan Boden, Al~Borchers, et~al.
\newblock In-datacenter performance analysis of a tensor processing unit.
\newblock In \emph{Proceedings of the 44th annual international symposium on computer architecture}, pp.\  1--12, 2017.

\bibitem[Judd et~al.(2016)Judd, Albericio, Hetherington, Aamodt, and Moshovos]{judd2016stripes}
Patrick Judd, Jorge Albericio, Tayler Hetherington, Tor~M Aamodt, and Andreas Moshovos.
\newblock Stripes: Bit-serial deep neural network computing.
\newblock In \emph{2016 49th Annual IEEE/ACM International Symposium on Microarchitecture (MICRO)}, pp.\  1--12. IEEE, 2016.

\bibitem[Kaplan et~al.(2020)Kaplan, McCandlish, Henighan, Brown, Chess, Child, Gray, Radford, Wu, and Amodei]{kaplan2020scaling}
Jared Kaplan, Sam McCandlish, Tom Henighan, Tom~B Brown, Benjamin Chess, Rewon Child, Scott Gray, Alec Radford, Jeffrey Wu, and Dario Amodei.
\newblock Scaling laws for neural language models.
\newblock \emph{arXiv preprint arXiv:2001.08361}, 2020.

\bibitem[Kenton \& Toutanova(2019)Kenton and Toutanova]{kenton2019bert}
Jacob Devlin Ming-Wei~Chang Kenton and Lee~Kristina Toutanova.
\newblock Bert: Pre-training of deep bidirectional transformers for language understanding.
\newblock In \emph{Proceedings of NAACL-HLT}, pp.\  4171--4186, 2019.

\bibitem[Killamsetty et~al.(2021{\natexlab{a}})Killamsetty, Durga, Ramakrishnan, De, and Iyer]{killamsetty2021grad}
Krishnateja Killamsetty, Sivasubramanian Durga, Ganesh Ramakrishnan, Abir De, and Rishabh Iyer.
\newblock Grad-match: Gradient matching based data subset selection for efficient deep model training.
\newblock In \emph{International Conference on Machine Learning}, pp.\  5464--5474. PMLR, 2021{\natexlab{a}}.

\bibitem[Killamsetty et~al.(2021{\natexlab{b}})Killamsetty, Sivasubramanian, Ramakrishnan, and Iyer]{killamsetty2021glister}
Krishnateja Killamsetty, Durga Sivasubramanian, Ganesh Ramakrishnan, and Rishabh Iyer.
\newblock Glister: Generalization based data subset selection for efficient and robust learning.
\newblock In \emph{Proceedings of the AAAI Conference on Artificial Intelligence}, volume~35, pp.\  8110--8118, 2021{\natexlab{b}}.

\bibitem[Killamsetty et~al.(2021{\natexlab{c}})Killamsetty, Zhao, Chen, and Iyer]{killamsetty2021retrieve}
Krishnateja Killamsetty, Xujiang Zhao, Feng Chen, and Rishabh Iyer.
\newblock Retrieve: Coreset selection for efficient and robust semi-supervised learning.
\newblock \emph{Advances in Neural Information Processing Systems}, 34:\penalty0 14488--14501, 2021{\natexlab{c}}.

\bibitem[Kim et~al.(2019)Kim, Bhalgat, Lee, Patel, and Kwak]{kim2019qkd}
Jangho Kim, Yash Bhalgat, Jinwon Lee, Chirag Patel, and Nojun Kwak.
\newblock Qkd: Quantization-aware knowledge distillation.
\newblock \emph{arXiv preprint arXiv:1911.12491}, 2019.

\bibitem[Kirillov et~al.(2023)Kirillov, Mintun, Ravi, Mao, Rolland, Gustafson, Xiao, Whitehead, Berg, Lo, et~al.]{kirillov2023segment}
Alexander Kirillov, Eric Mintun, Nikhila Ravi, Hanzi Mao, Chloe Rolland, Laura Gustafson, Tete Xiao, Spencer Whitehead, Alexander~C Berg, Wan-Yen Lo, et~al.
\newblock Segment anything.
\newblock \emph{arXiv preprint arXiv:2304.02643}, 2023.

\bibitem[Kirsch(2023)]{kirsch2023does}
Andreas Kirsch.
\newblock Does ‘deep learning on a data diet’reproduce? overall yes, but grand at initialization does not.
\newblock \emph{Transactions on Machine Learning Research}, 2023.

\bibitem[Krizhevsky et~al.(2009)Krizhevsky, Hinton, et~al.]{krizhevsky2009learning}
Alex Krizhevsky, Geoffrey Hinton, et~al.
\newblock Learning multiple layers of features from tiny images.
\newblock 2009.

\bibitem[Krizhevsky et~al.(2012)Krizhevsky, Sutskever, and Hinton]{krizhevsky2012imagenet}
Alex Krizhevsky, Ilya Sutskever, and Geoffrey~E Hinton.
\newblock Imagenet classification with deep convolutional neural networks.
\newblock \emph{Advances in neural information processing systems}, 25:\penalty0 1097--1105, 2012.

\bibitem[Lee et~al.(2023)Lee, Kim, Kwon, and Lee]{FlexRound}
Jung~Hyun Lee, Jeonghoon Kim, Se~Jung Kwon, and Dongsoo Lee.
\newblock Flexround: Learnable rounding based on element-wise division for post-training quantization.
\newblock In \emph{International Conference on Machine Learning}, pp.\  18913--18939. PMLR, 2023.

\bibitem[Lee et~al.(2019)Lee, Yun, Lee, Lee, Li, and Shin]{lee2019robust}
Kimin Lee, Sukmin Yun, Kibok Lee, Honglak Lee, Bo~Li, and Jinwoo Shin.
\newblock Robust inference via generative classifiers for handling noisy labels.
\newblock In \emph{International conference on machine learning}, pp.\  3763--3772. PMLR, 2019.

\bibitem[Li et~al.(2018)Li, Xu, Taylor, Studer, and Goldstein]{li2018visualizing}
Hao Li, Zheng Xu, Gavin Taylor, Christoph Studer, and Tom Goldstein.
\newblock Visualizing the loss landscape of neural nets.
\newblock In \emph{Proceedings of the 32nd International Conference on Neural Information Processing Systems}, pp.\  6391--6401, 2018.

\bibitem[Li et~al.(2024)Li, Xiao, Shi, Zhang, Yang, Liu, and Chen]{li2024nicest}
Lin Li, Jun Xiao, Hanrong Shi, Hanwang Zhang, Yi~Yang, Wei Liu, and Long Chen.
\newblock Nicest: Noisy label correction and training for robust scene graph generation.
\newblock \emph{IEEE Transactions on Pattern Analysis and Machine Intelligence}, 2024.

\bibitem[Li et~al.(2019{\natexlab{a}})Li, Wang, Liang, Qin, Yan, and Fan]{li2019fully}
Rundong Li, Yan Wang, Feng Liang, Hongwei Qin, Junjie Yan, and Rui Fan.
\newblock Fully quantized network for object detection.
\newblock In \emph{Proceedings of the IEEE/CVF conference on computer vision and pattern recognition}, pp.\  2810--2819, 2019{\natexlab{a}}.

\bibitem[Li et~al.(2023)Li, Han, Shan, and Chen]{li2023disc}
Yifan Li, Hu~Han, Shiguang Shan, and Xilin Chen.
\newblock Disc: Learning from noisy labels via dynamic instance-specific selection and correction.
\newblock In \emph{Proceedings of the IEEE/CVF Conference on Computer Vision and Pattern Recognition}, pp.\  24070--24079, 2023.

\bibitem[Li et~al.(2019{\natexlab{b}})Li, Dong, and Wang]{li2019apot}
Yuhang Li, Xin Dong, and Wei Wang.
\newblock Additive powers-of-two quantization: An efficient non-uniform discretization for neural networks.
\newblock In \emph{International Conference on Learning Representations}, 2019{\natexlab{b}}.

\bibitem[Lin et~al.(2014)Lin, Maire, Belongie, Hays, Perona, Ramanan, Doll{\'a}r, and Zitnick]{lin2014microsoft}
Tsung-Yi Lin, Michael Maire, Serge Belongie, James Hays, Pietro Perona, Deva Ramanan, Piotr Doll{\'a}r, and C~Lawrence Zitnick.
\newblock Microsoft coco: Common objects in context.
\newblock In \emph{Computer Vision--ECCV 2014: 13th European Conference, Zurich, Switzerland, September 6-12, 2014, Proceedings, Part V 13}, pp.\  740--755. Springer, 2014.

\bibitem[Lin et~al.(2017)Lin, Goyal, Girshick, He, and Doll{\'a}r]{lin2017focal}
Tsung-Yi Lin, Priya Goyal, Ross Girshick, Kaiming He, and Piotr Doll{\'a}r.
\newblock Focal loss for dense object detection.
\newblock In \emph{Proceedings of the IEEE international conference on computer vision}, pp.\  2980--2988, 2017.

\bibitem[Liu et~al.(2023{\natexlab{a}})Liu, Liu, and Cheng]{liu2023oscillation}
Shih-Yang Liu, Zechun Liu, and Kwang-Ting Cheng.
\newblock Oscillation-free quantization for low-bit vision transformers.
\newblock \emph{arXiv preprint arXiv:2302.02210}, 2023{\natexlab{a}}.

\bibitem[Liu et~al.(2023{\natexlab{b}})Liu, Liu, Huang, Dong, and Cheng]{liu2023llm}
Shih-yang Liu, Zechun Liu, Xijie Huang, Pingcheng Dong, and Kwang-Ting Cheng.
\newblock Llm-fp4: 4-bit floating-point quantized transformers.
\newblock In \emph{The 2023 Conference on Empirical Methods in Natural Language Processing}, 2023{\natexlab{b}}.

\bibitem[Liu et~al.(2019)Liu, Mu, Zhang, Guo, Yang, Cheng, and Sun]{liu2019metapruning}
Zechun Liu, Haoyuan Mu, Xiangyu Zhang, Zichao Guo, Xin Yang, Kwang-Ting Cheng, and Jian Sun.
\newblock Metapruning: Meta learning for automatic neural network channel pruning.
\newblock In \emph{Proceedings of the IEEE/CVF international conference on computer vision}, pp.\  3296--3305, 2019.

\bibitem[Liu et~al.(2022)Liu, Cheng, Huang, Xing, and Shen]{liu2022nonuniform}
Zechun Liu, Kwang-Ting Cheng, Dong Huang, Eric~P Xing, and Zhiqiang Shen.
\newblock Nonuniform-to-uniform quantization: Towards accurate quantization via generalized straight-through estimation.
\newblock In \emph{Proceedings of the IEEE/CVF Conference on Computer Vision and Pattern Recognition}, pp.\  4942--4952, 2022.

\bibitem[Liu et~al.(2023{\natexlab{c}})Liu, Oguz, Zhao, Chang, Stock, Mehdad, Shi, Krishnamoorthi, and Chandra]{liu2023llmqat}
Zechun Liu, Barlas Oguz, Changsheng Zhao, Ernie Chang, Pierre Stock, Yashar Mehdad, Yangyang Shi, Raghuraman Krishnamoorthi, and Vikas Chandra.
\newblock Llm-qat: Data-free quantization aware training for large language models.
\newblock \emph{arXiv preprint arXiv:2305.17888}, 2023{\natexlab{c}}.

\bibitem[Liu et~al.(2017)Liu, Li, Shen, Huang, Yan, and Zhang]{liu2017learning}
Zhuang Liu, Jianguo Li, Zhiqiang Shen, Gao Huang, Shoumeng Yan, and Changshui Zhang.
\newblock Learning efficient convolutional networks through network slimming.
\newblock In \emph{Proceedings of the IEEE international conference on computer vision}, pp.\  2736--2744, 2017.

\bibitem[Liu et~al.(2018)Liu, Sun, Zhou, Huang, and Darrell]{liu2018rethinking}
Zhuang Liu, Mingjie Sun, Tinghui Zhou, Gao Huang, and Trevor Darrell.
\newblock Rethinking the value of network pruning.
\newblock In \emph{International Conference on Learning Representations}, 2018.

\bibitem[Lopez-Paz et~al.(2015)Lopez-Paz, Bottou, Sch{\"o}lkopf, and Vapnik]{lopez2015unifying}
David Lopez-Paz, L{\'e}on Bottou, Bernhard Sch{\"o}lkopf, and Vladimir Vapnik.
\newblock Unifying distillation and privileged information.
\newblock \emph{arXiv preprint arXiv:1511.03643}, 2015.

\bibitem[Margatina et~al.(2021)Margatina, Vernikos, Barrault, and Aletras]{margatina2021active}
Katerina Margatina, Giorgos Vernikos, Lo{\"\i}c Barrault, and Nikolaos Aletras.
\newblock Active learning by acquiring contrastive examples.
\newblock In \emph{Proceedings of the 2021 Conference on Empirical Methods in Natural Language Processing}, pp.\  650--663, 2021.

\bibitem[Minoux(1978)]{minoux1978accelerated}
Michel Minoux.
\newblock Accelerated greedy algorithms for maximizing submodular set functions.
\newblock In \emph{Optimization techniques}, pp.\  234--243. Springer, 1978.

\bibitem[Mirzadeh et~al.(2020)Mirzadeh, Farajtabar, Li, Levine, Matsukawa, and Ghasemzadeh]{mirzadeh2020improved}
Seyed~Iman Mirzadeh, Mehrdad Farajtabar, Ang Li, Nir Levine, Akihiro Matsukawa, and Hassan Ghasemzadeh.
\newblock Improved knowledge distillation via teacher assistant.
\newblock In \emph{Proceedings of the AAAI conference on artificial intelligence}, volume~34, pp.\  5191--5198, 2020.

\bibitem[Mirzasoleiman et~al.(2013)Mirzasoleiman, Karbasi, Sarkar, and Krause]{mirzasoleiman2013distributed}
Baharan Mirzasoleiman, Amin Karbasi, Rik Sarkar, and Andreas Krause.
\newblock Distributed submodular maximization: Identifying representative elements in massive data.
\newblock In \emph{Advances in Neural Information Processing Systems}, pp.\  2049--2057, 2013.

\bibitem[Mirzasoleiman et~al.(2020)Mirzasoleiman, Bilmes, and Leskovec]{mirzasoleiman2020coresets}
Baharan Mirzasoleiman, Jeff Bilmes, and Jure Leskovec.
\newblock Coresets for data-efficient training of machine learning models.
\newblock In \emph{International Conference on Machine Learning}, pp.\  6950--6960. PMLR, 2020.

\bibitem[Mishra \& Marr(2018)Mishra and Marr]{mishra2018apprentice}
Asit Mishra and Debbie Marr.
\newblock Apprentice: Using knowledge distillation techniques to improve low-precision network accuracy.
\newblock In \emph{International Conference on Learning Representations}, 2018.

\bibitem[Miyashita et~al.(2016)Miyashita, Lee, and Murmann]{miyashita2016convolutional}
Daisuke Miyashita, Edward~H Lee, and Boris Murmann.
\newblock Convolutional neural networks using logarithmic data representation.
\newblock \emph{arXiv preprint arXiv:1603.01025}, 2016.

\bibitem[Molchanov et~al.(2019)Molchanov, Mallya, Tyree, Frosio, and Kautz]{molchanov2019importance}
Pavlo Molchanov, Arun Mallya, Stephen Tyree, Iuri Frosio, and Jan Kautz.
\newblock Importance estimation for neural network pruning.
\newblock In \emph{Proceedings of the IEEE/CVF Conference on Computer Vision and Pattern Recognition}, pp.\  11264--11272, 2019.

\bibitem[Nagel et~al.(2020)Nagel, Amjad, Van~Baalen, Louizos, and Blankevoort]{nagel2020up}
Markus Nagel, Rana~Ali Amjad, Mart Van~Baalen, Christos Louizos, and Tijmen Blankevoort.
\newblock Up or down? adaptive rounding for post-training quantization.
\newblock In \emph{International Conference on Machine Learning}, pp.\  7197--7206. PMLR, 2020.

\bibitem[Nemhauser et~al.(1978)Nemhauser, Wolsey, and Fisher]{nemhauser1978analysis}
George~L Nemhauser, Laurence~A Wolsey, and Marshall~L Fisher.
\newblock An analysis of approximations for maximizing submodular set functions—i.
\newblock \emph{Mathematical programming}, 14\penalty0 (1):\penalty0 265--294, 1978.

\bibitem[Park et~al.(2024)Park, Choi, Kim, Song, and Lee]{park2024robust}
Dongmin Park, Seola Choi, Doyoung Kim, Hwanjun Song, and Jae-Gil Lee.
\newblock Robust data pruning under label noise via maximizing re-labeling accuracy.
\newblock \emph{Advances in Neural Information Processing Systems}, 36, 2024.

\bibitem[Park et~al.(2019)Park, Kim, Lu, and Cho]{park2019relational}
Wonpyo Park, Dongju Kim, Yan Lu, and Minsu Cho.
\newblock Relational knowledge distillation.
\newblock In \emph{Proceedings of the IEEE/CVF Conference on Computer Vision and Pattern Recognition}, pp.\  3967--3976, 2019.

\bibitem[Paul et~al.(2021)Paul, Ganguli, and Dziugaite]{paul2021el2n}
Mansheej Paul, Surya Ganguli, and Gintare~Karolina Dziugaite.
\newblock Deep learning on a data diet: Finding important examples early in training.
\newblock \emph{Advances in Neural Information Processing Systems}, 34:\penalty0 20596--20607, 2021.

\bibitem[Pham et~al.(2018)Pham, Guan, Zoph, Le, and Dean]{pham2018efficient}
Hieu Pham, Melody Guan, Barret Zoph, Quoc Le, and Jeff Dean.
\newblock Efficient neural architecture search via parameters sharing.
\newblock In \emph{International Conference on Machine Learning}, pp.\  4095--4104. PMLR, 2018.

\bibitem[Polino et~al.()Polino, Pascanu, and Alistarh]{polinomodel}
Antonio Polino, Razvan Pascanu, and Dan Alistarh.
\newblock Model compression via distillation and quantization.
\newblock In \emph{International Conference on Learning Representations}.

\bibitem[Pooladzandi et~al.(2022)Pooladzandi, Davini, and Mirzasoleiman]{pooladzandi2022adaptive}
Omead Pooladzandi, David Davini, and Baharan Mirzasoleiman.
\newblock Adaptive second order coresets for data-efficient machine learning.
\newblock In \emph{International Conference on Machine Learning}, pp.\  17848--17869. PMLR, 2022.

\bibitem[Sener \& Savarese(2018)Sener and Savarese]{seneractive}
Ozan Sener and Silvio Savarese.
\newblock Active learning for convolutional neural networks: A core-set approach.
\newblock In \emph{International Conference on Learning Representations}, 2018.

\bibitem[Sharma et~al.(2018)Sharma, Park, Suda, Lai, Chau, Chandra, and Esmaeilzadeh]{sharma2018bit}
Hardik Sharma, Jongse Park, Naveen Suda, Liangzhen Lai, Benson Chau, Vikas Chandra, and Hadi Esmaeilzadeh.
\newblock Bit fusion: Bit-level dynamically composable architecture for accelerating deep neural network.
\newblock In \emph{2018 ACM/IEEE 45th Annual International Symposium on Computer Architecture (ISCA)}, pp.\  764--775. IEEE, 2018.

\bibitem[Shen \& Xing(2021)Shen and Xing]{shen2021fast}
Zhiqiang Shen and Eric Xing.
\newblock A fast knowledge distillation framework for visual recognition.
\newblock \emph{arXiv preprint arXiv:2112.01528}, 2021.

\bibitem[Song et~al.(2022)Song, Kim, Park, Shin, and Lee]{song2022learning}
Hwanjun Song, Minseok Kim, Dongmin Park, Yooju Shin, and Jae-Gil Lee.
\newblock Learning from noisy labels with deep neural networks: A survey.
\newblock \emph{IEEE transactions on neural networks and learning systems}, 2022.

\bibitem[Sorscher et~al.(2022)Sorscher, Geirhos, Shekhar, Ganguli, and Morcos]{sorscher2022beyond}
Ben Sorscher, Robert Geirhos, Shashank Shekhar, Surya Ganguli, and Ari Morcos.
\newblock Beyond neural scaling laws: beating power law scaling via data pruning.
\newblock \emph{Advances in Neural Information Processing Systems}, 35:\penalty0 19523--19536, 2022.

\bibitem[Tan \& Le(2019)Tan and Le]{tan2019efficientnet}
Mingxing Tan and Quoc Le.
\newblock Efficientnet: Rethinking model scaling for convolutional neural networks.
\newblock In \emph{International Conference on Machine Learning}, pp.\  6105--6114. PMLR, 2019.

\bibitem[Toneva et~al.(2019)Toneva, Sordoni, des Combes, Trischler, Bengio, and Gordon]{toneva2018forgetting}
Mariya Toneva, Alessandro Sordoni, Remi~Tachet des Combes, Adam Trischler, Yoshua Bengio, and Geoffrey~J. Gordon.
\newblock An empirical study of example forgetting during deep neural network learning.
\newblock In \emph{International Conference on Learning Representations}, 2019.

\bibitem[Touvron et~al.(2023)Touvron, Martin, Stone, Albert, Almahairi, Babaei, Bashlykov, Batra, Bhargava, Bhosale, et~al.]{touvron2023llama}
Hugo Touvron, Louis Martin, Kevin Stone, Peter Albert, Amjad Almahairi, Yasmine Babaei, Nikolay Bashlykov, Soumya Batra, Prajjwal Bhargava, Shruti Bhosale, et~al.
\newblock Llama 2: Open foundation and fine-tuned chat models.
\newblock \emph{arXiv preprint arXiv:2307.09288}, 2023.

\bibitem[Tu et~al.(2023)Tu, Zhang, Li, Liu, Li, Wang, Wang, and Zhao]{tu2023learning}
Yuanpeng Tu, Boshen Zhang, Yuxi Li, Liang Liu, Jian Li, Yabiao Wang, Chengjie Wang, and Cai~Rong Zhao.
\newblock Learning from noisy labels with decoupled meta label purifier.
\newblock In \emph{Proceedings of the IEEE/CVF Conference on Computer Vision and Pattern Recognition}, pp.\  19934--19943, 2023.

\bibitem[Vapnik(1999)]{vapnik1999overview}
Vladimir~N Vapnik.
\newblock An overview of statistical learning theory.
\newblock \emph{IEEE transactions on neural networks}, 10\penalty0 (5):\penalty0 988--999, 1999.

\bibitem[Wang et~al.(2020)Wang, Chen, He, and Cheng]{wang2020towards}
Peisong Wang, Qiang Chen, Xiangyu He, and Jian Cheng.
\newblock Towards accurate post-training network quantization via bit-split and stitching.
\newblock In \emph{International Conference on Machine Learning}, pp.\  9847--9856. PMLR, 2020.

\bibitem[Wei et~al.(2021)Wei, Tao, Xie, and An]{wei2021open}
Hongxin Wei, Lue Tao, Renchunzi Xie, and Bo~An.
\newblock Open-set label noise can improve robustness against inherent label noise.
\newblock \emph{Advances in Neural Information Processing Systems}, 34:\penalty0 7978--7992, 2021.

\bibitem[Welling(2009)]{welling2009herding}
Max Welling.
\newblock Herding dynamical weights to learn.
\newblock In \emph{Proceedings of the 26th Annual International Conference on Machine Learning}, pp.\  1121--1128, 2009.

\bibitem[Wolf(2011)]{wolf2011facility}
Gert~W Wolf.
\newblock Facility location: concepts, models, algorithms and case studies. series: Contributions to management science, 2011.

\bibitem[Xia et~al.(2023{\natexlab{a}})Xia, Han, Zhan, Yu, Gong, Gong, and Liu]{xia2023combating}
Xiaobo Xia, Bo~Han, Yibing Zhan, Jun Yu, Mingming Gong, Chen Gong, and Tongliang Liu.
\newblock Combating noisy labels with sample selection by mining high-discrepancy examples.
\newblock In \emph{Proceedings of the IEEE/CVF International Conference on Computer Vision}, pp.\  1833--1843, 2023{\natexlab{a}}.

\bibitem[Xia et~al.(2023{\natexlab{b}})Xia, Liu, Yu, Shen, Han, and Liu]{xia2023moderate}
Xiaobo Xia, Jiale Liu, Jun Yu, Xu~Shen, Bo~Han, and Tongliang Liu.
\newblock Moderate coreset: A universal method of data selection for real-world data-efficient deep learning.
\newblock In \emph{The Eleventh International Conference on Learning Representations}, 2023{\natexlab{b}}.

\bibitem[Xiao et~al.(2023)Xiao, Lin, Seznec, Wu, Demouth, and Han]{xiao2023smoothquant}
Guangxuan Xiao, Ji~Lin, Mickael Seznec, Hao Wu, Julien Demouth, and Song Han.
\newblock Smoothquant: Accurate and efficient post-training quantization for large language models.
\newblock In \emph{International Conference on Machine Learning}, pp.\  38087--38099. PMLR, 2023.

\bibitem[Xiao et~al.(2012)Xiao, Xiao, and Eckert]{xiao2012adversarial}
Han Xiao, Huang Xiao, and Claudia Eckert.
\newblock Adversarial label flips attack on support vector machines.
\newblock In \emph{ECAI 2012}, pp.\  870--875. IOS Press, 2012.

\bibitem[Yang et~al.(2019)Yang, Dai, Yang, Carbonell, Salakhutdinov, and Le]{yang2019xlnet}
Zhilin Yang, Zihang Dai, Yiming Yang, Jaime Carbonell, Russ~R Salakhutdinov, and Quoc~V Le.
\newblock Xlnet: Generalized autoregressive pretraining for language understanding.
\newblock \emph{Advances in neural information processing systems}, 32, 2019.

\bibitem[Yao et~al.(2018)Yao, Wang, Tsang, Zhang, Sun, Zhang, and Zhang]{yao2018deep}
Jiangchao Yao, Jiajie Wang, Ivor~W Tsang, Ya~Zhang, Jun Sun, Chengqi Zhang, and Rui Zhang.
\newblock Deep learning from noisy image labels with quality embedding.
\newblock \emph{IEEE Transactions on Image Processing}, 28\penalty0 (4):\penalty0 1909--1922, 2018.

\bibitem[Yvinec et~al.(2023)Yvinec, Dapogny, Cord, and Bailly]{yvinec2023powerquant}
Edouard Yvinec, Arnaud Dapogny, Matthieu Cord, and Kevin Bailly.
\newblock Powerquant: Automorphism search for non-uniform quantization.
\newblock In \emph{The Eleventh International Conference on Learning Representations}, 2023.

\bibitem[Zhang et~al.(2018)Zhang, Yang, Ye, and Hua]{zhang2018lqnet}
Dongqing Zhang, Jiaolong Yang, Dongqiangzi Ye, and Gang Hua.
\newblock Lq-nets: Learned quantization for highly accurate and compact deep neural networks.
\newblock In \emph{Proceedings of the European conference on computer vision (ECCV)}, pp.\  365--382, 2018.

\bibitem[Zhou et~al.(2016)Zhou, Wu, Ni, Zhou, Wen, and Zou]{zhou2016dorefa}
Shuchang Zhou, Yuxin Wu, Zekun Ni, Xinyu Zhou, He~Wen, and Yuheng Zou.
\newblock Dorefa-net: Training low bitwidth convolutional neural networks with low bitwidth gradients.
\newblock \emph{arXiv preprint arXiv:1606.06160}, 2016.

\bibitem[Zhou et~al.(2022)Zhou, Pi, Zhang, Lin, Chen, and Zhang]{zhou2022probabilistic}
Xiao Zhou, Renjie Pi, Weizhong Zhang, Yong Lin, Zonghao Chen, and Tong Zhang.
\newblock Probabilistic bilevel coreset selection.
\newblock In \emph{International Conference on Machine Learning}, pp.\  27287--27302. PMLR, 2022.

\bibitem[Zhuang et~al.(2020)Zhuang, Liu, Tan, Shen, and Reid]{zhuang2020training}
Bohan Zhuang, Lingqiao Liu, Mingkui Tan, Chunhua Shen, and Ian Reid.
\newblock Training quantized neural networks with a full-precision auxiliary module.
\newblock In \emph{Proceedings of the IEEE/CVF conference on computer vision and pattern recognition}, pp.\  1488--1497, 2020.

\end{thebibliography}
\bibliographystyle{tmlr}

\newpage
\appendix
\section*{Appendix}
This appendix includes an additional introduction to related works, training dynamics, efficiency analysis, extended experimental analysis on data coverage, and a discussion of transferability and generalizability that is not included in the main text due to space limitations. These contents are organized in separate sections as follows:
\begin{itemize}
    \item Sec.~\ref{sec:baseline} elaborates the detail of previous representative coreset selection methods and analyzes the reason for failure in quantization-aware training (QAT) scenario.
    
    \item Sec.~\ref{sec:dynamic} demonstrate the training dynamics of our ACS with different selection intervals and show how our methods help to stabilize the training and accelerate the convergence.

    \item Sec.~\ref{sec:efficiency} includes the real training time composition as a supplement of QAT training time comparison in Tab.~\ref{tab:efficiency}, showing that our coreset selection only incurs a minimal efficiency overhead both with and without knowledge distillation.

    \item Sec.~\ref{sec:coverage} provides additional experiments to prove that the improvement of performance with our method does not exclusively come from covering more data from the full training set.

    \item Sec.~\ref{sec:kd} provides detailed experimental results with and without knowledge distillation (KD). Both the accuracy and real training time are shown.

    \item Sec.~\ref{sec:overlap} analyzes the transferability and generalizability of our coreset from different models.

\end{itemize}

\section{Detailed Introduction of Selected Baselines}\label{sec:baseline}

In this section, we will introduce the selected baseline coreset selection methods in detail and show the defects of these methods when applied to QAT. 
\paragraph{EL2N-Score/GraNd-Score~\citep{paul2021el2n}} The GraNd-score for a given training sample $\{x,y\}$ is defined as $\chi_{t}(x,y) = E_{w_t} \left\|g_{t} (x,y) \right\|_2$, which is the expected magnitude of the loss vector with respect to the weights. The importance of each sample is measured by the expected loss gradient norm, which has a similar intuition to our error-vector score $d_{\text{EVS}}$. However, another assumption from GraNd score is that this approximation only holds when the model has been trained for a few epochs. We must perform early training on the current model to have statistics to compute the score. In addition, storing all the gradients incurs significant memory overheads during QAT. The efficiency will be lower than full dataset training with a high subset fraction. The performance with these metrics is sub-optimal as the converged quantized model in the later training epochs of QAT is not considered. In addition, Based on the reproduction from \citet{kirsch2023does} observing that the inconsistency of GraNd at initialization, the performance of GraNd-score-based coreset selection cannot outperform random selection across various datasets and tasks.

\paragraph{Forgetting~\citep{toneva2018forgetting}} The core contribution of Forgetting is that forgetting and learning events are defined. In the classification setting with a given dataset $\mathcal{D} = (\mathbf{x}_i, y_i)_i$. For training example $\mathbf{x}_i$ obtained after $t$ steps using SGD, the predicted label is denoted as $\hat y_i^t = \arg\max_k p(y_{ik} |\mathbf{x}_i; \theta^{t})$. $\text{acc}_i^t = 1_{\hat{y}_i^t=y_i}$ is defined to be a binary indicator encoding whether the classification of the specific example is correct at time step $t$. Example $\mathbf{x}_i$ undergoes a \emph{forgetting event} when $\text{acc}_i^t$ decreases between two consecutive updates: $\text{acc}_i^t > \text{acc}_i^{t+1}$. The example $\mathbf{x}_i$ is misclassified at step $t+1$ after having been correctly classified at step $t$. Corresponding to the forgetting event, a~\emph{learning event} is defined to be $\text{acc}_i^t < \text{acc}_i^{t + 1}$. With the statistics of forgetting events in early training, we can select those samples that incur more forgetting events and are more difficult to learn. However, the intuition to select ``difficult'' samples that incur more misclassification is not always correct. For large-scale datasets (ImageNet-1K, etc.) and datasets with a relatively smaller size (such as MNIST, CIFAR-10, etc.). the difficulties of classification vary significantly. Selecting samples that are difficult to learn at the very beginning is not always reasonable. In the quantization-aware training setting, the quantized model will first calibrate the quantization parameters and recover the weight at the early stages. We should not select ``difficult samples'' for calibration and recovery. 

\paragraph{Glister~\citep{killamsetty2021glister}} The Glister performs data selection based on the assumption that the inner discrete data selection is an instance of (weakly) submodular optimization. Let $V = \{1, 2, \cdots, n\}$ denote a ground set of items (the set of training samples in the setting of coreset selection). Set functions are defined as functions $f: 2^V \rightarrow \mathbf{R}$ that operate on subsets of $V$. A set function $f$ is defined to be a submodular function~\citep{fujishige2005submodular} if it satisfies the diminishing returns property that for subsets $S \subseteq T \subseteq V, f(j | S) \triangleq  f(S \cup j) - f(S) \geq f(j | T)$. Some natural combinatorial functions (facility location, set cover, concave over modularity, etc.) are submodular functions~\citep{iyer2020submodular}. Submodularity is also very appealing because a simple greedy algorithm achieves a $1 - 1/e$ constant factor approximation guarantee~\citep{nemhauser1978analysis} for the problem of maximizing a submodular function subject to a cardinality constraint (which most data selection approaches involve). Moreover, several variants of the greedy algorithm have been proposed, further scaling up submodular maximization to almost linear time complexity~\citep{minoux1978accelerated,mirzasoleiman2013distributed}. However, the greedy algorithm still consumes tremendous time compared to other metric-based selection methods, which only need to perform sorting on the proposed metrics. We also observed that the greedy algorithm fails on QAT for low-bit settings.

\paragraph{kCenterGreedy~\citep{seneractive}} As one of the most straight-forward geometry-based coreset selection methods, the intuition is simple: we can measure the similarity by the distance of data points and select the center point of data cluster can make use of the redundancy in the dataset. This method aims to solve the \textit{minimax facility location} problem \citep{wolf2011facility}, which is defined to be selecting $k$ samples as subset $S$ from the full dataset $T$ such that the longest distance between a data point in $T \backslash S$ and its closest data point in $S$ is minimized:
\begin{equation}
    \min_{S\subset T} \max_{x_i\in T\backslash S} \min_{x_j\in S} \mathcal{D}(x_i,x_j),
\end{equation}
where $\mathcal{D}(\cdot,\cdot)$ is the distance measurement function. The problem is NP-hard, and a greedy approximation known as \textsc{k-Center Greedy} has been proposed in \citet{seneractive}. Similarly, as the greedy algorithm is involved, the efficiency is negatively influenced, which is not affordable for our QAT setting.

\paragraph{Contextual Diversity (CD)~\citep{agarwal2020contextual}} Specifically designed for coreset selection for deep convolutional neural networks (CNNs), Contextual Diversity (CD) fully leverages the ambiguity in feature representations. The ambiguity is quantified as class-specific confusion as the selection metric. 
Assume $ C = \{1,\ldots,n_C \} $ is the set of classes predicted by a CNN-based model. For a region $r$ within an input image $I$, let $  P_r = P_r(\widehat{y} \mid I ; \theta)$ be the softmax probability vector as predicted by the model $ \theta $. The pseudo-label for the region $ r \subseteq I $ is defined as $\widehat{y}_r = \arg\max_{j\in C} P_r[j]$, where the notation $ P_r[j] $ denotes the $ j^{th} $ element of the vector. For a given model $ \theta $ over the unlabeled $I$, the class-specific confusion for class $c$ is defined as $ P_{\tiny I}^c = \frac{1}{|I^c|} \sum_{I\in I^c} \left[ \frac{\sum_{\tiny r \in R_{\tiny I}^c} w_{r} P_r(\widehat{y} \mid I; \theta)}{\sum_{\tiny r\in R_{\tiny I}^c}{w_r}} \right]$ with $ w_r \geq 0$ as the mixing weights. The pairwise contextual diversity, which is the KL-divergence of the metric between two samples $I_1$ and $I_2$ could be used as a distance metric to replace the Euclidean distance in the previous kCenterGreedy~\citep{seneractive}. As this work basically follows kCenterGreedy~\citep{seneractive} to perform coreset selection, the defects also lie in the efficiency of the greedy algorithm and failure with the low-bit setting.

\paragraph{Moderate Coreset~\citep{xia2023moderate}} As the most recent work of coreset selection, Moderate has optimal efficiency as no greedy searching or back-propagation is involved during selection. Previous methods rely on score criterion that only applies to a specific scenario. This work proposes to utilize the score median as a proxy of the statistical score distribution and select the data points with scores close to the score median into the coreset. The main drawback of this method applied to quantization-aware training is that quantized distribution is not considered.

\section{QAT Training Dynamics with ACS} \label{sec:dynamic}

\begin{figure}[h]
  \centering
  \begin{subfigure}{0.24\linewidth}
    \includegraphics[width=\textwidth]{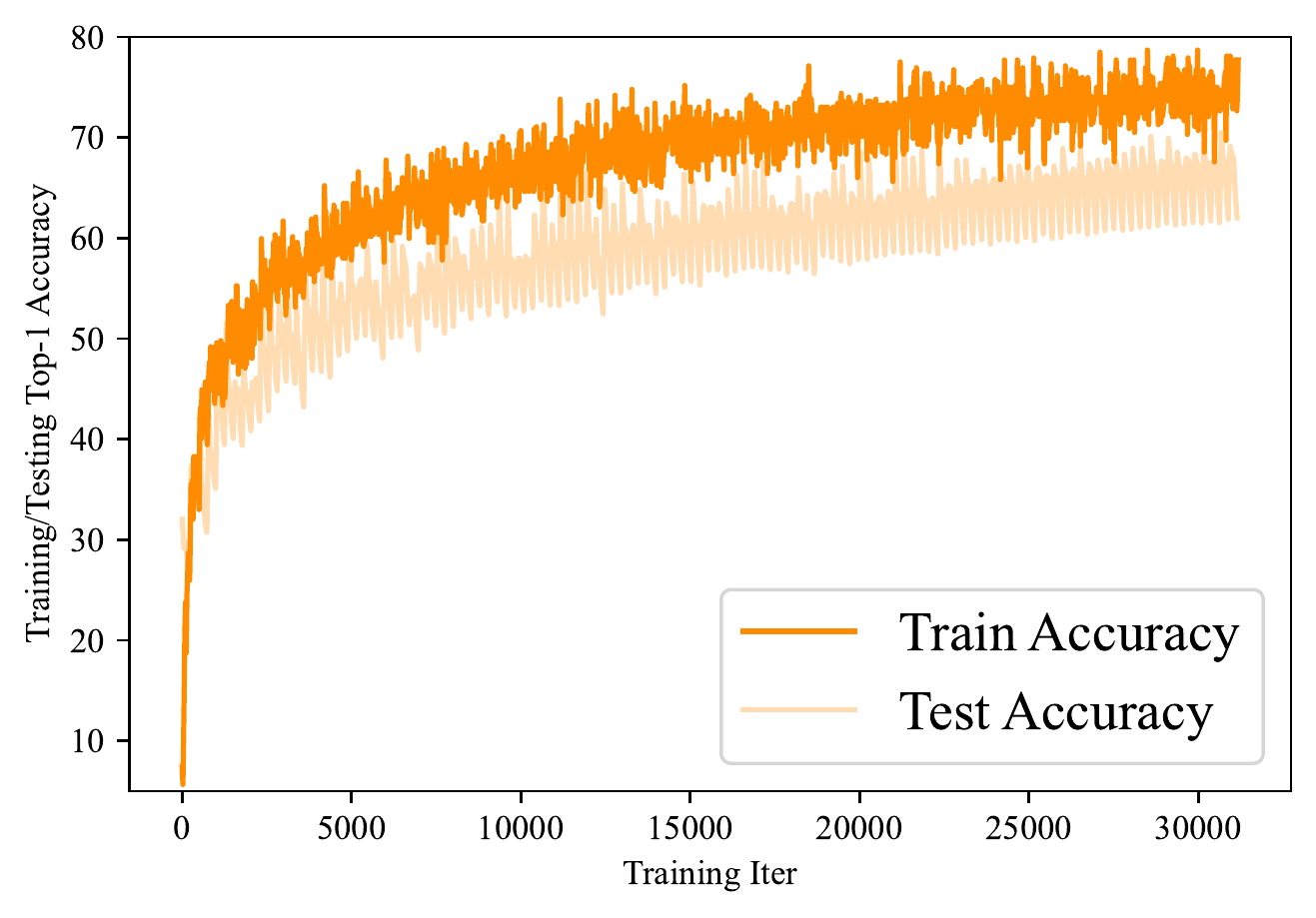}
    \caption{Acc-random}
  \end{subfigure}
  \begin{subfigure}{0.24\linewidth}
    \includegraphics[width=\textwidth]{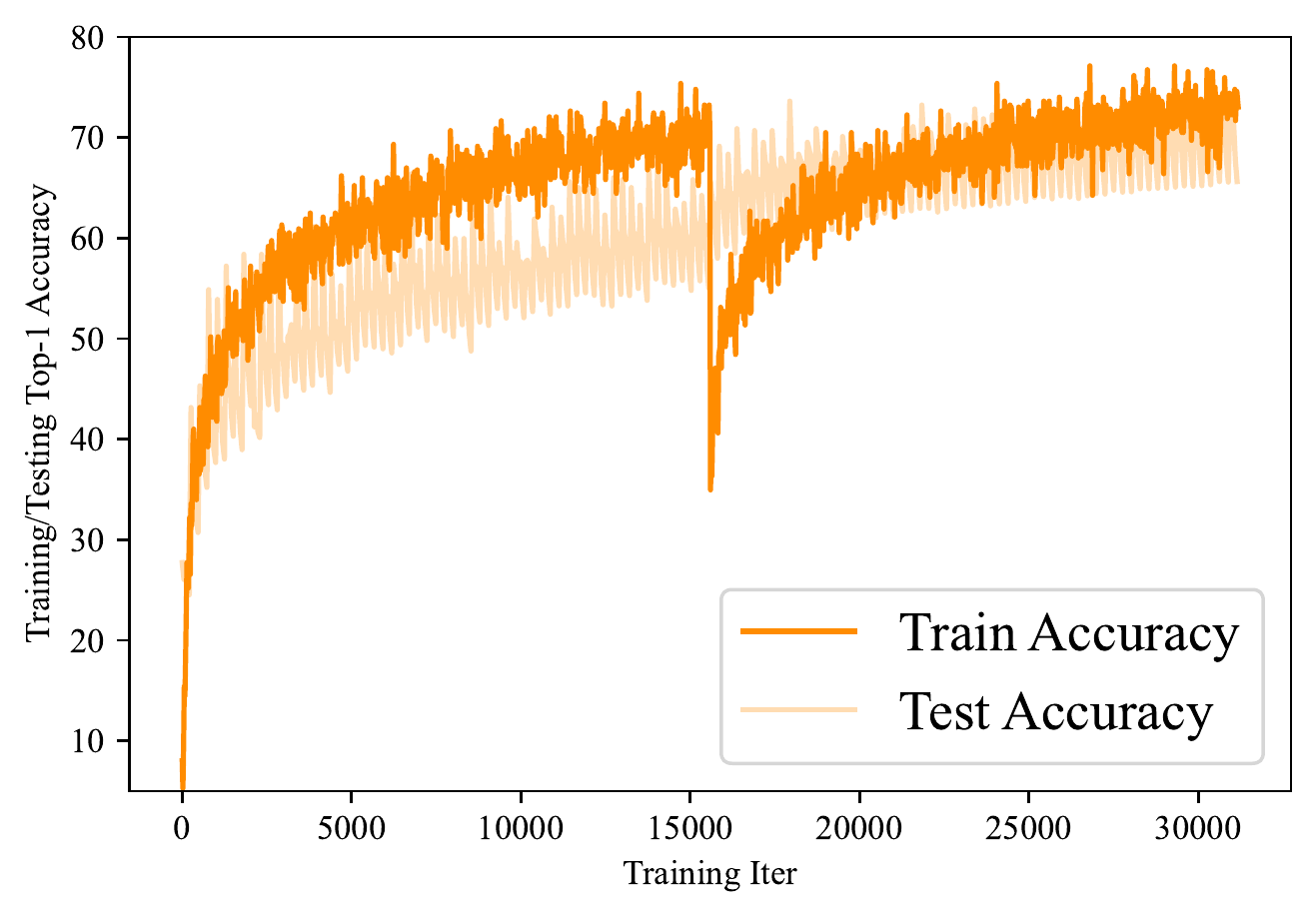}
    \caption{R=60}
  \end{subfigure}
  \begin{subfigure}{0.24\linewidth}
    \includegraphics[width=\textwidth]{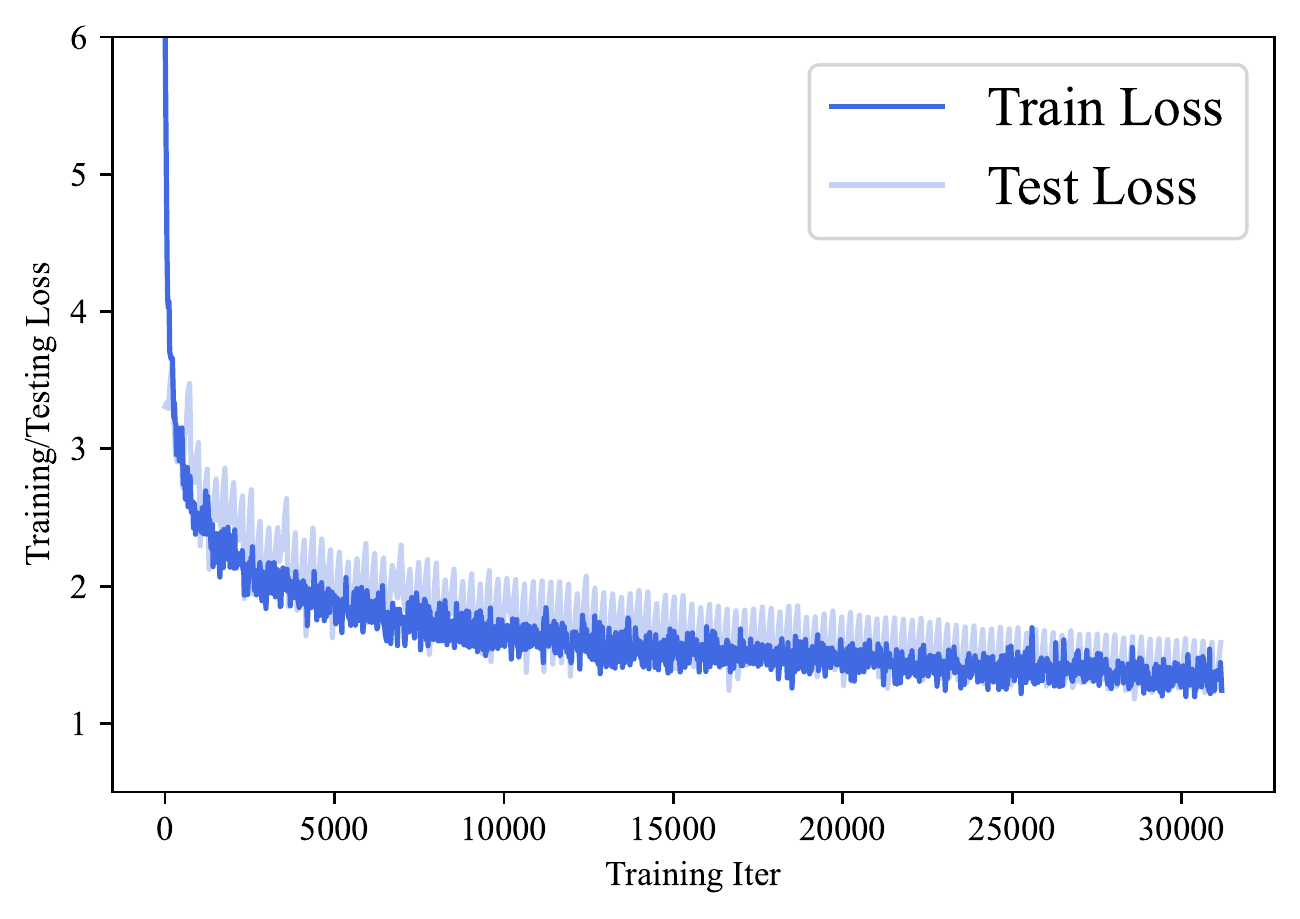}
    \caption{Loss-random}
  \end{subfigure}
  \begin{subfigure}{0.24\linewidth}
    \includegraphics[width=\textwidth]{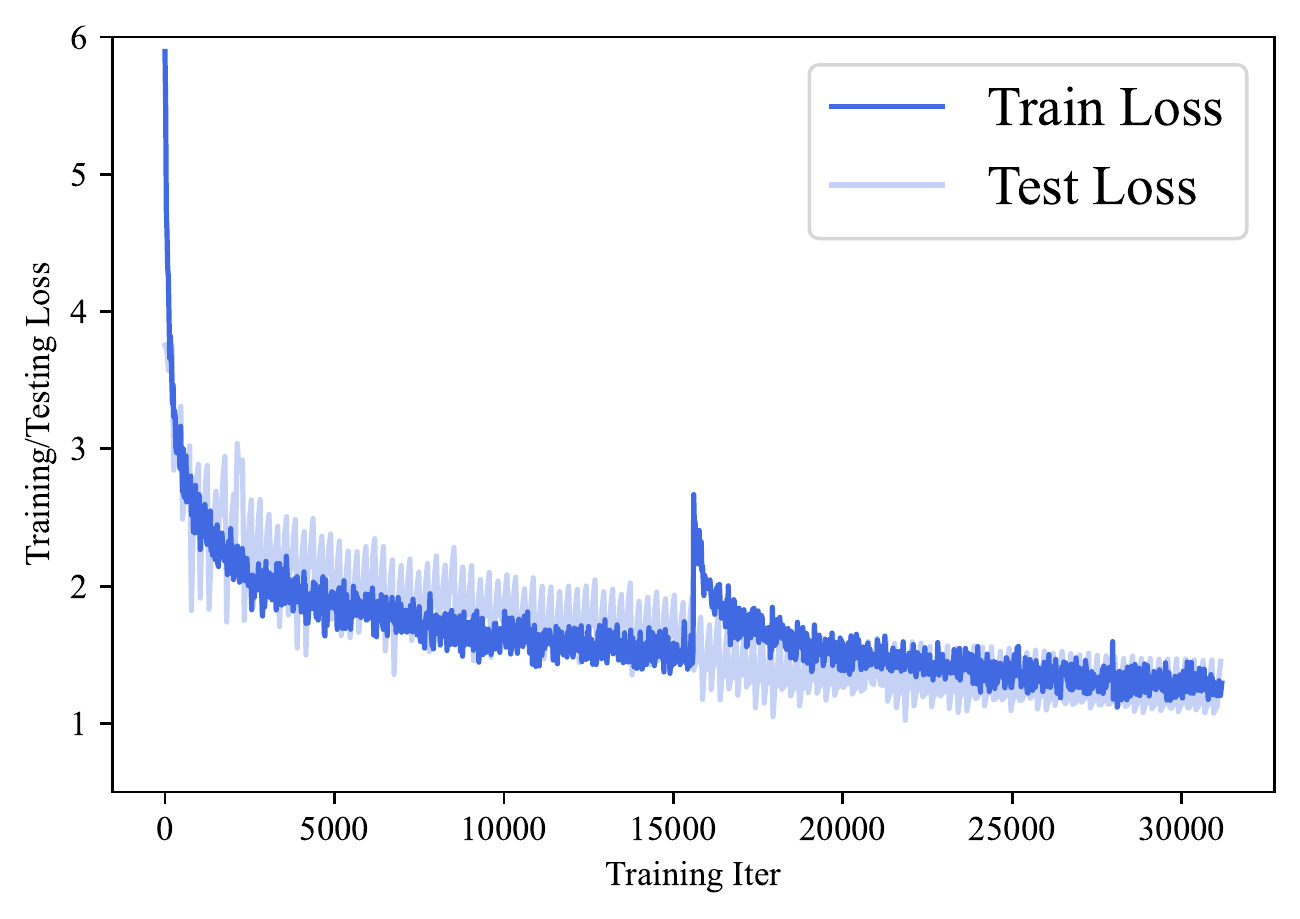}
    \caption{R=60}
  \end{subfigure}
  \begin{subfigure}{0.24\linewidth}
    \includegraphics[width=\textwidth]{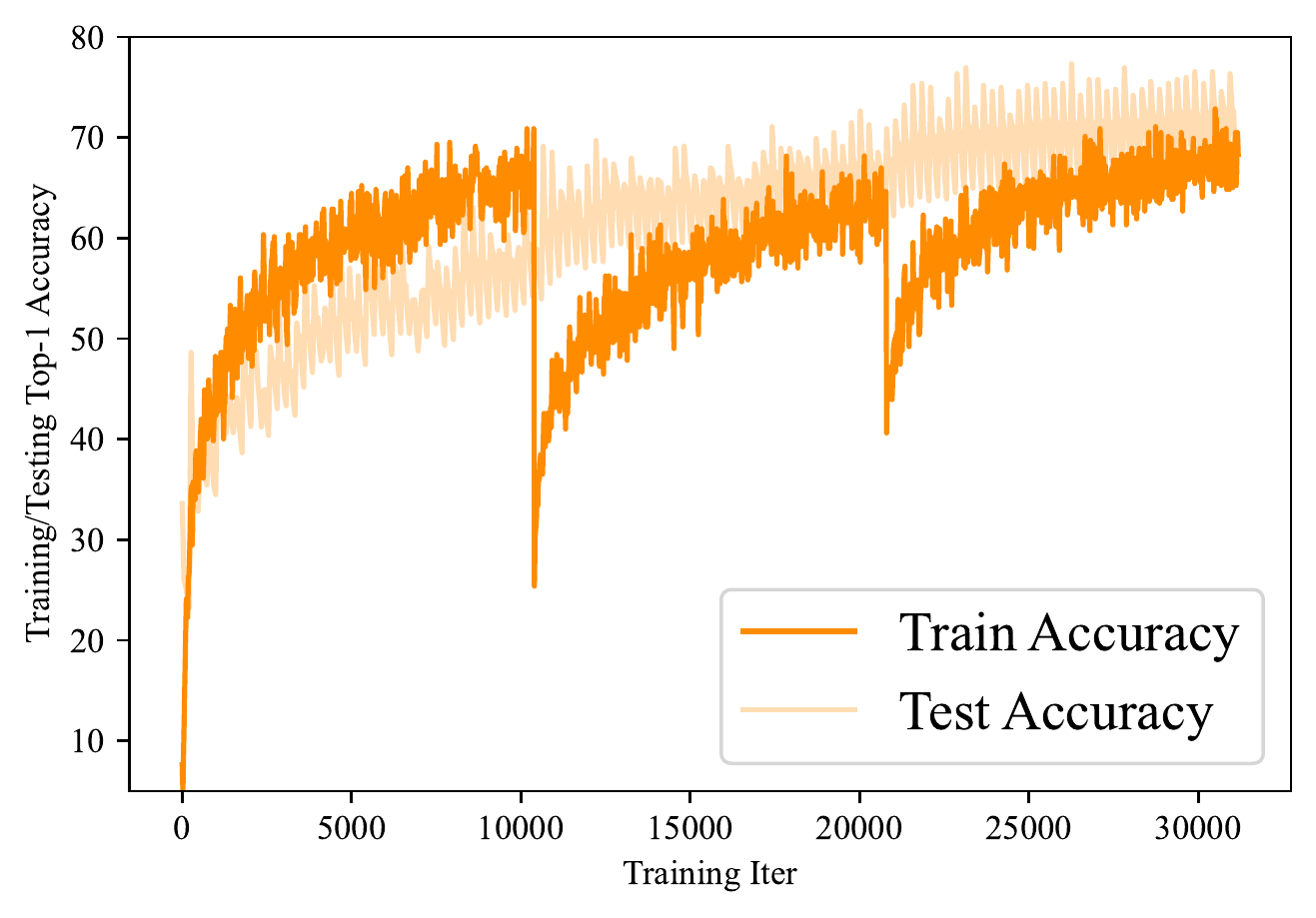}
    \caption{R=40}
  \end{subfigure}
  \begin{subfigure}{0.24\linewidth}
    \includegraphics[width=\textwidth]{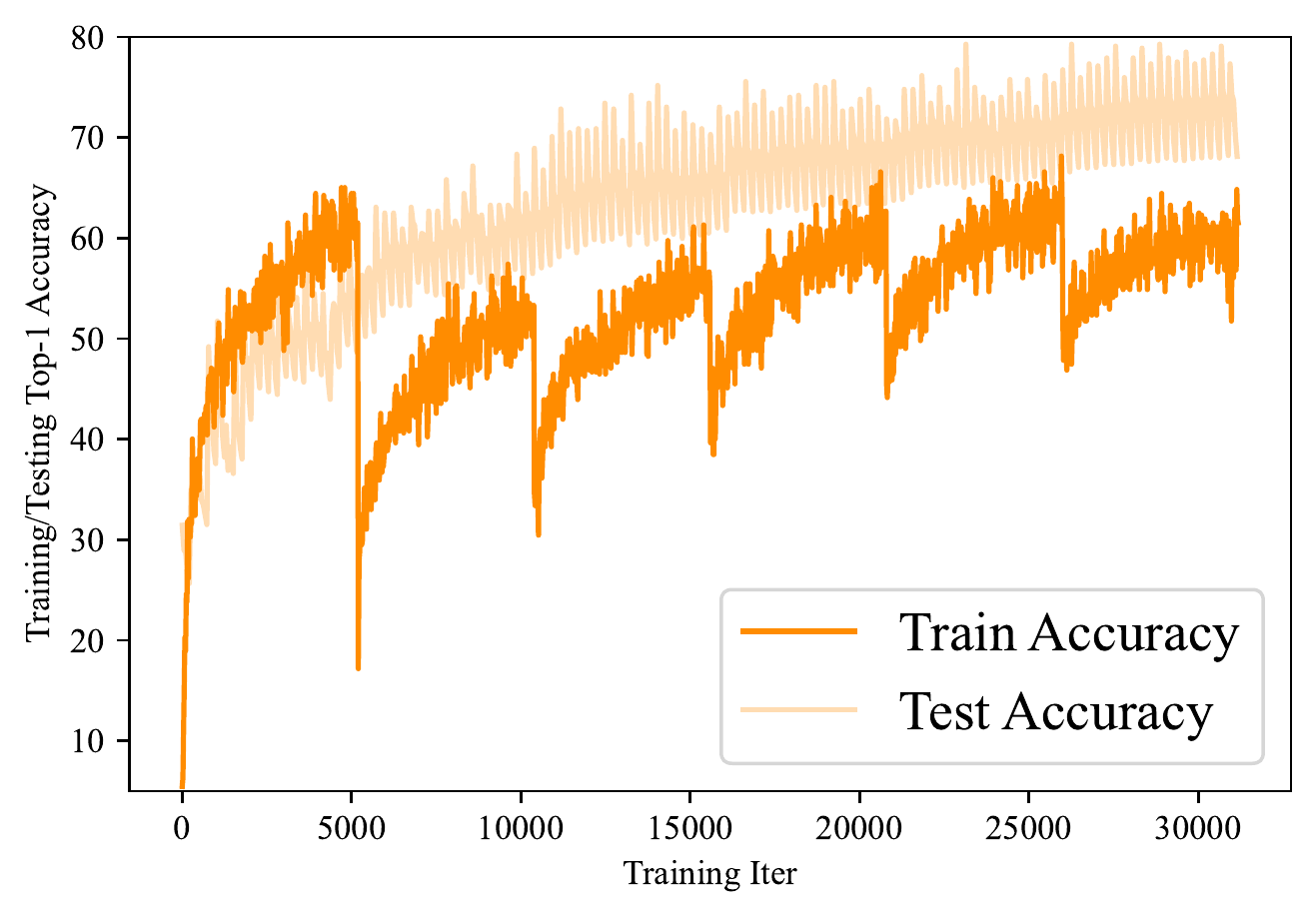}
    \caption{R=20}
  \end{subfigure}
  \begin{subfigure}{0.24\linewidth}
    \includegraphics[width=\textwidth]{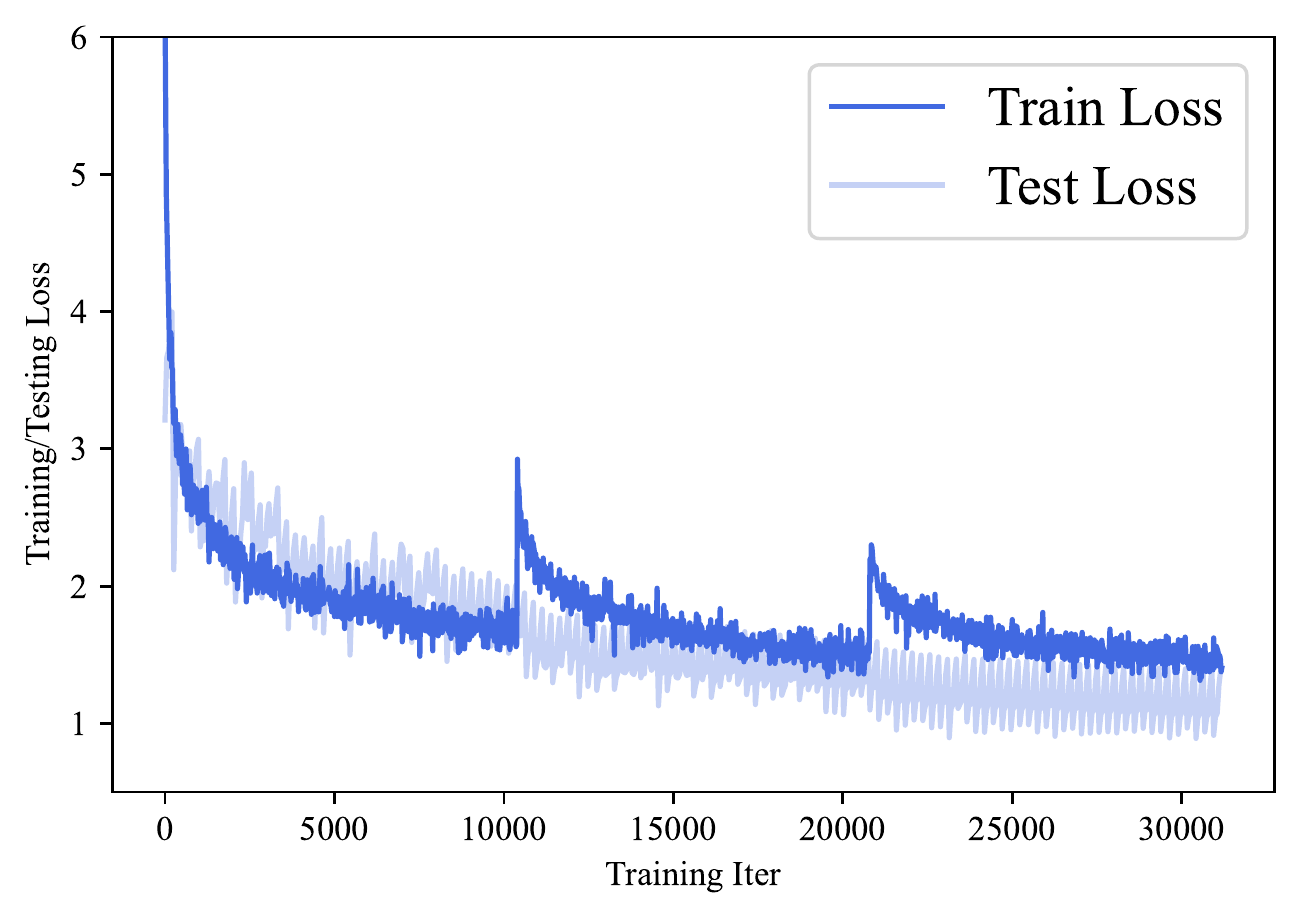}
    \caption{R=40}
  \end{subfigure}
  \begin{subfigure}{0.24\linewidth}
    \includegraphics[width=\textwidth]{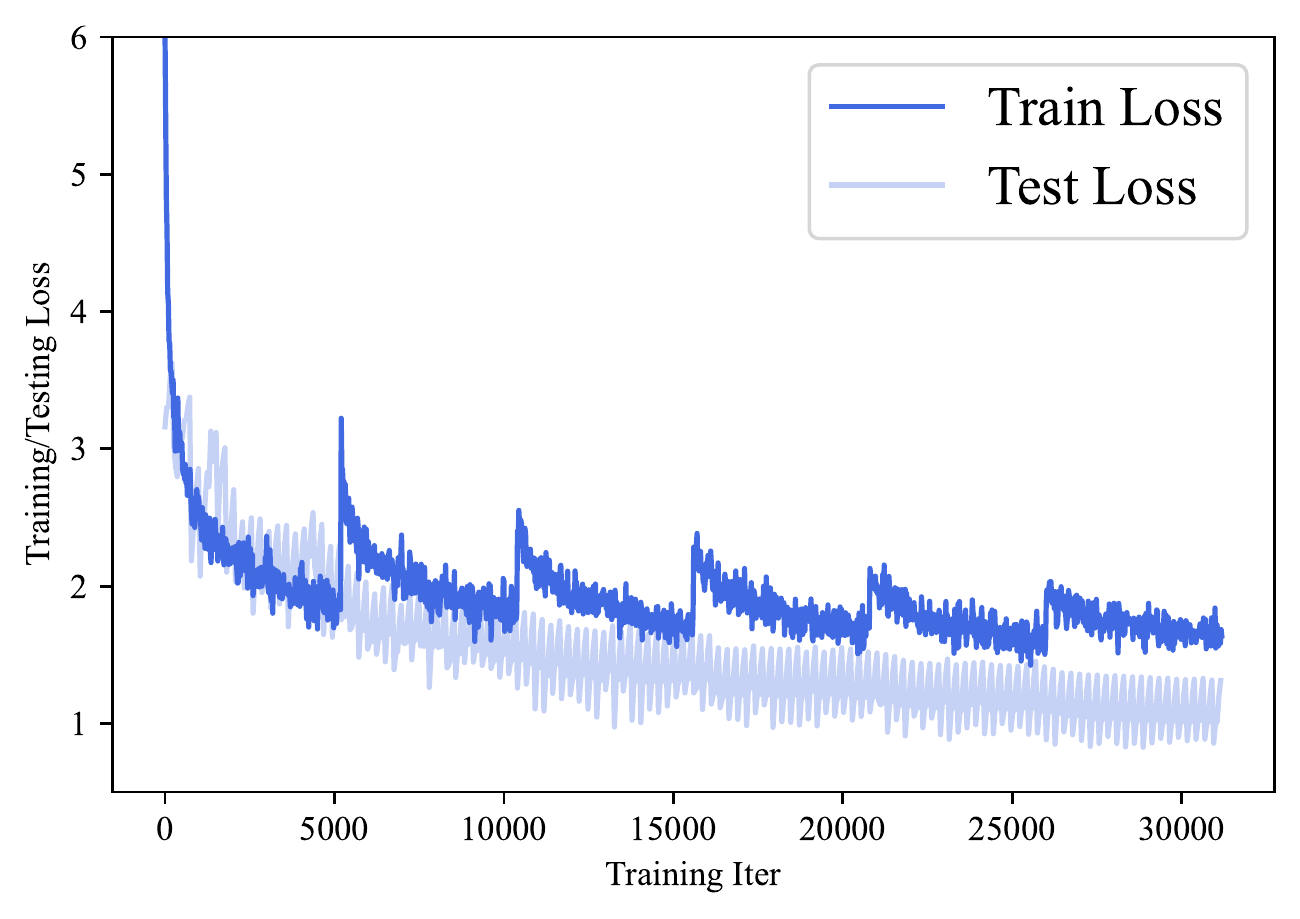}
    \caption{R=20}
  \end{subfigure}
  \begin{subfigure}{0.24\linewidth}
    \includegraphics[width=\textwidth]{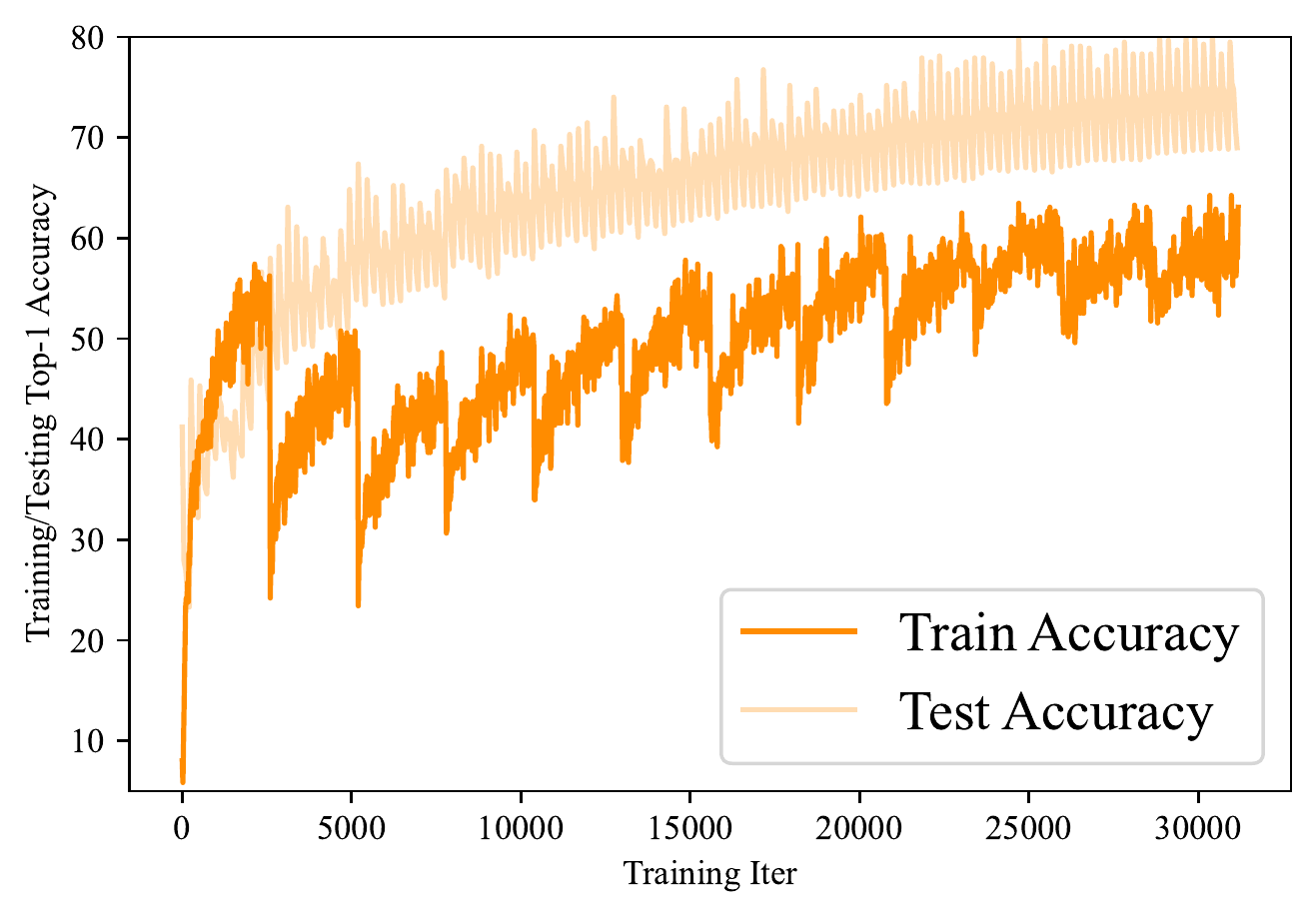}
    \caption{R=10}
  \end{subfigure}
  \begin{subfigure}{0.24\linewidth}
    \includegraphics[width=\textwidth]{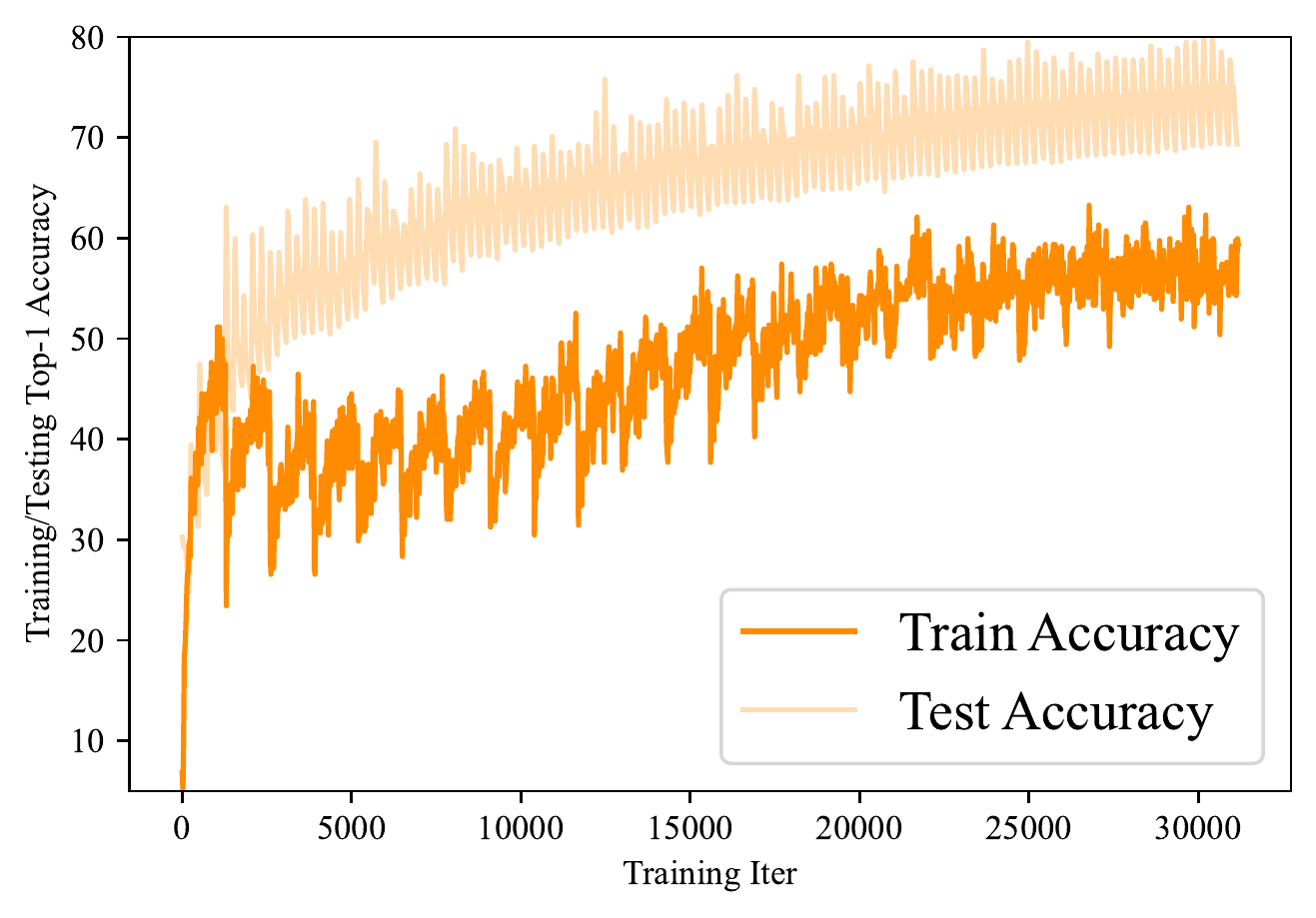}
    \caption{R=5}
  \end{subfigure}
  \begin{subfigure}{0.24\linewidth}
    \includegraphics[width=\textwidth]{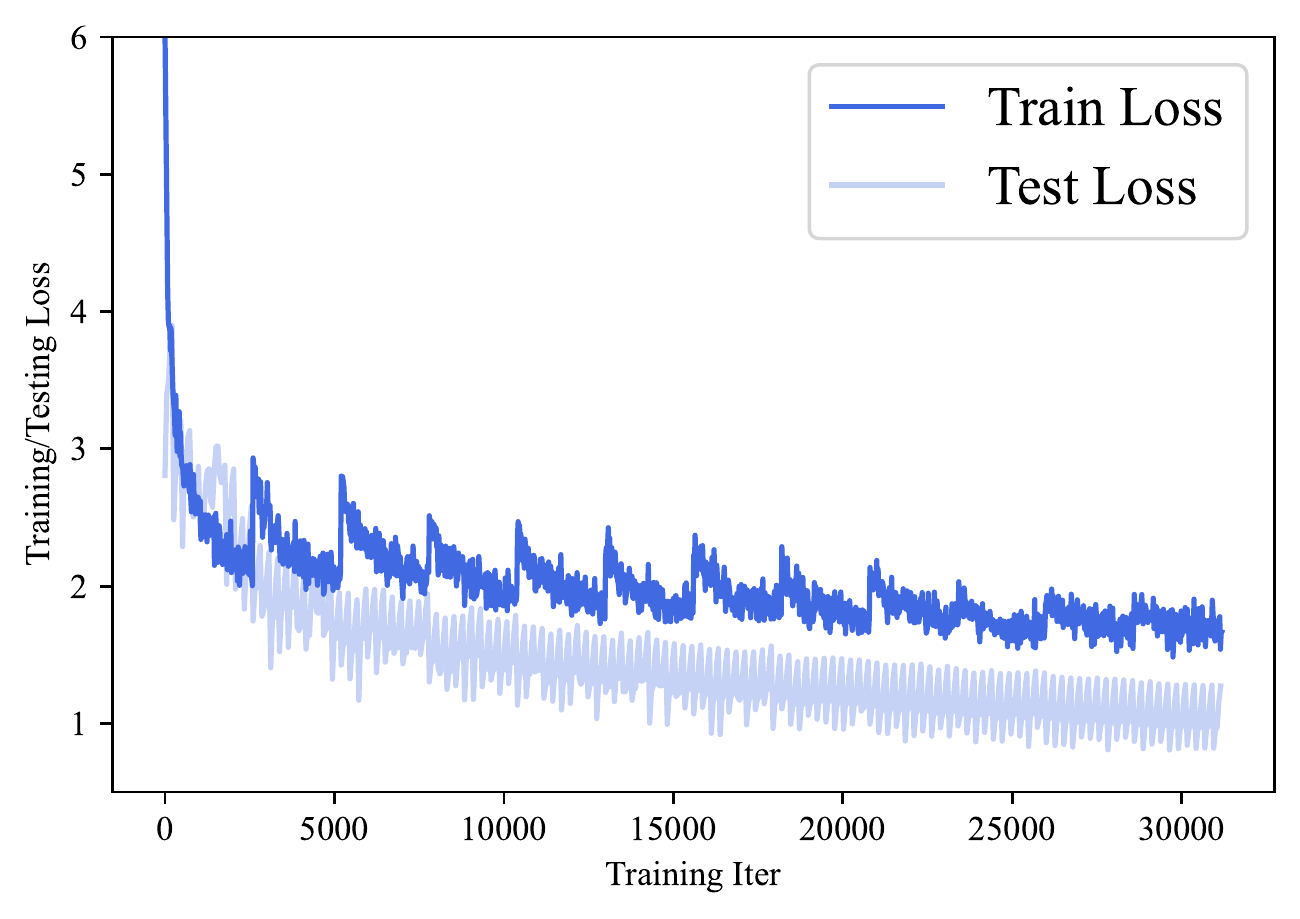}
    \caption{R=10}
  \end{subfigure}
  \begin{subfigure}{0.24\linewidth}
    \includegraphics[width=\textwidth]{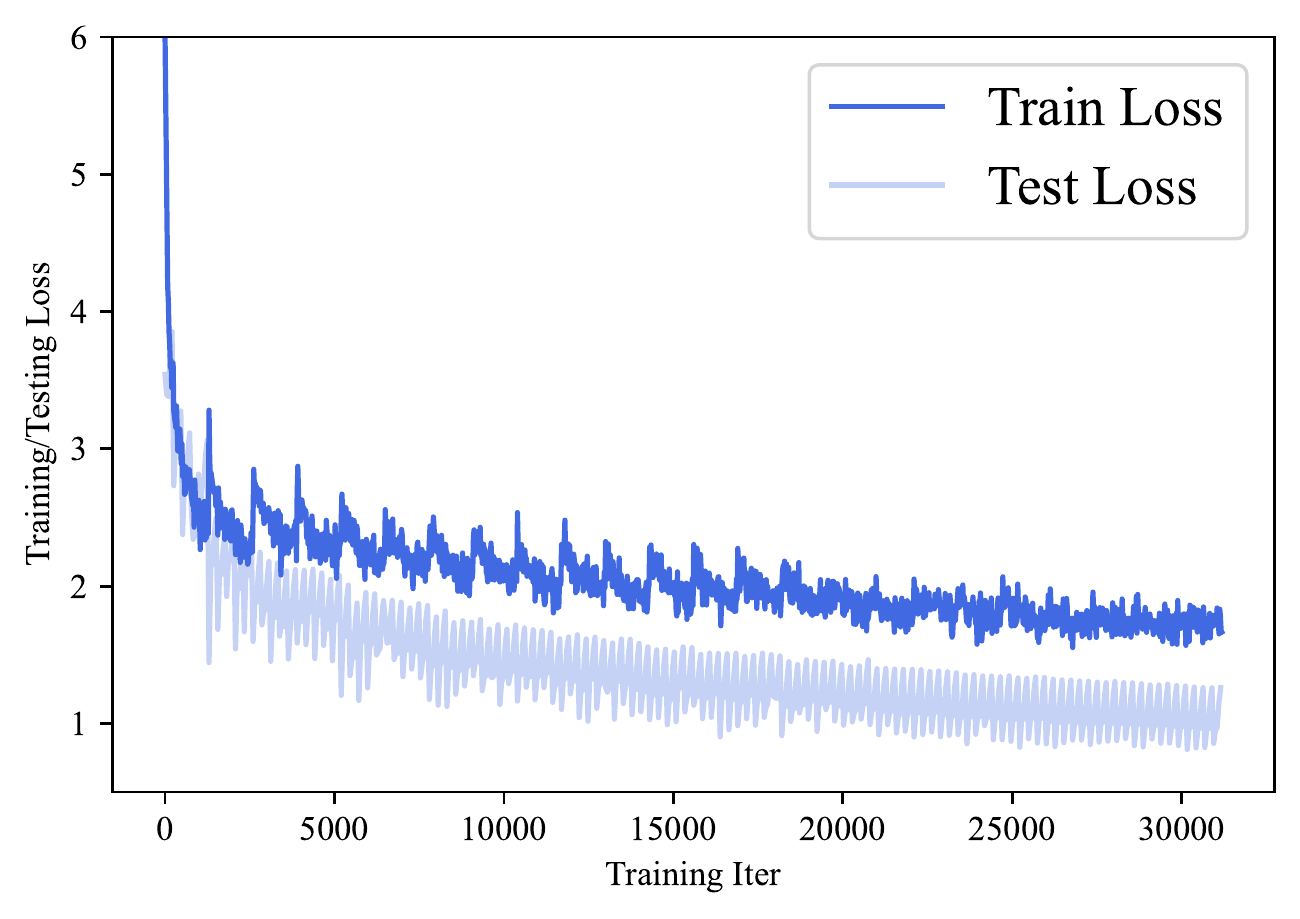}
    \caption{R=5}
  \end{subfigure}

  \caption{Training/Testing loss/accuracy comparison of 4/4-bit quantized ResNet-18 on 10\% coreset of ImageNet-1K with different selection interval R.}
  \label{fig:dynamic}

\end{figure}

In this section, we report the training dynamics, including the training loss and training accuracy of QAT ResNet-18 on ImageNet-1K coreset with a 10\% subset fraction. As can be seen from the results in Fig.~\ref{fig:dynamic}, when the coreset changes adaptively to the current iterations, the accuracy will drop, and loss will increase significantly at the specific iteration of updating the subset. However, the quantized model will converge fast on the new subset. The training with an adaptive coreset can effectively help avoid overfitting and improve the final performance.

\section{Training Time Composition} \label{sec:efficiency}

As a supplement of the detailed QAT training time comparison in Tab.~\ref{tab:efficiency}, we also provide the detailed training time composition of training and selection in Tab.~\ref{tab:efficiency2}. Since KD will change the loss function of SGD training and is difficult to break down these two parts, we provide two sets of results with and without KD. The experiments are also conducted on quantized ResNet-18 on ImageNet-1K coreset with 2 NVIDIA RTX 3090 GPUs.

The results show that coreset selection only incurs a minimal efficiency overhead both with and without KD. The selection time is a constant and is only related to the full dataset size and the selection intervals $R$. The training with KD and without KD only influence the backpropagation time of the training and there is no backward in the selection phase, which makes the time of selection the same (3.2h) across all settings.

\begin{table}[h]
  \caption{Composition of training time on QAT of quantized ResNet-18 on ImageNet-1K with different subset fractions and KD settings. The bitwidth for quantized ResNet-18 is 4/4 for weights/activations.}
  \label{tab:efficiency2}
  \centering
  \resizebox{0.7\textwidth}{!}{
  \begin{tabular}{lccccccc}
    \toprule
    Stage/Fraction (\%) & Apply KD?  & 10\% &30\% &50\% &60\% &70\% &80\%  \\
    \midrule
    Ours-all       & \checkmark & 11.3h &20.7h & 36.1h & 42.2h & 48.0h & 53.8h\\
    Ours-selection & \checkmark & 3.2h & 3.2h & 3.2h & 3.2h & 3.2h & 3.2h\\
    Ours-training  & \checkmark & 8.1h & 17.5h & 32.9h & 39.0h & 44.8h & 50.6h\\
    \midrule
    Ours-all       &  & 9.7h & 16.0h & 29.5h & 34.4h & 40.2h & 45.4h\\
    Ours-selection &  & 3.2h & 3.2h & 3.2h & 3.2h & 3.2h & 3.2h\\
    Ours-training  &  & 6.5h & 15.8h & 26.3h & 31.2h & 37.0h & 42.2h\\
    \bottomrule
  \end{tabular}}
\end{table}

\section{Coreset Coverage Analysis} \label{sec:coverage}

The advantages of the proposed quantization-aware Adaptive Coreset Selection (ACS) algorithms are two-fold: adaption and diversity. In this section, we further demonstrate that improving diversity by covering all the training data in different subsets is not optimal compared to our adaptive coreset. We use the ``Full Coverage Split'' of the CIFAR-100 dataset as our baseline, which is to select non-overlapped samples randomly into the subset of fraction $S$ in the first $\lceil 1/S \rceil$ epochs. It is guaranteed that all the training samples are included, but the sequence is random in this setting. We apply QAT to a 2/32-bit quantized MobileNet-V2 on the random subset, ``Full Coverage Split'' subset, and coreset using our methods. The selection interval $R$ is the same across all settings. When the subset fraction $S$ is large, the difference in coverage rate with various methods is minor. We then only evaluate on the $S \in \{10\%, 20\%, 30\%, 40\%, 50\%\}$
The results are shown in Tab.~\ref{tab:coverage}, where we can see that ``Full Coverage Split'' has limited superiority on performance compared to random baseline, especially when the subset fraction R is large. Our method outperforms the other two settings across all fractions, proving that both adaption and diversity help to improve the performance.

\begin{table}[h]
  \caption{Comparision of Top-1 Accuracy with Random vs. Full coverage split vs. Ours}
  \label{tab:coverage}
  \centering
  \begin{tabular}{lcccccc}
    \toprule
    Method/Fraction (\%)   & 10\% &20\% &30\% &40\% &50\%\\
    \midrule
    Random                 &62.25$\pm$0.71 &64.07$\pm$0.47 &65.22$\pm$0.41 &65.55$\pm$0.80 & 66.24$\pm$0.55   \\
    Full Coverage Split    &62.55$\pm$0.65 & 64.22$\pm$0.81 & 65.34$\pm$1.17 &65.69$\pm$0.69 & 66.20$\pm$0.89\\
    \midrule
    Ours    &\textbf{63.37$\pm$0.55}&\textbf{65.91$\pm$0.40}&\textbf{66.41$\pm$0.31}&\textbf{66.86$\pm$0.72}&\textbf{67.19$\pm$0.65} \\
    \bottomrule
  \end{tabular}
\end{table}

\section{Detailed Experimental Results without Knowledge Distillation}\label{sec:kd}

Knowledge Distillation (KD) is a normal and ``default'’ setting for all previous quantization work~\citep{mishra2018apprentice,zhuang2020training,liu2022nonuniform,bhalgat2020lsq+,huang2022sdq}, including the LSQ quantization~\citep{bhalgat2020lsq+} we use in this paper. For a fair comparison with previous work, we equally utilize KD for all the experiments in this work regardless of the precision and dataset fraction. The full-data training baseline also involves training with knowledge distillation.

To verify the effectiveness of our method without KD, we remove the knowledge distillation in our method. Since the DS is built on the soft label of the teacher model, we do not use it in our selection w/o KD. Only EVS is applied as the selection metric. We follow the same settings for quantized ResNet-18 on ImageNet-1K as shown in our paper. The training time and accuracy are shown as follows in Tab.~\ref{tab:nokd}. When full data are selected ($S=100\%$), the Top-1 accuracy is 70.21\% and training time is 3.1h without KD. From the results shown in Tab.~\ref{tab:nokd}, we can see that our method still outperforms previous SOTA without KD and the training efficiency is still optimal.

\begin{table}[h]
  \caption{Comparison of Top-1 Accuracy and the training time of QAT of quantized ResNet-18 on ImageNet-1K with different subset fractions without KD settings. The bitwidth for quantized ResNet-18 is 4/4 for weights/activations.}
  \label{tab:nokd}
  \centering
  \resizebox{0.98\textwidth}{!}{
  \begin{tabular}{lccccccccccccc}
    \toprule
    \multirow{2}{*}{Method/Fraction (\%)}   & \multicolumn{2}{c}{10\%} &\multicolumn{2}{c}{30\%} & \multicolumn{2}{c}{50\%} & \multicolumn{2}{c}{60\%} & \multicolumn{2}{c}{70\%} & \multicolumn{2}{c}{80\%}  \\
    &Acc&Time&Acc&Time&Acc&Time&Acc&Time&Acc&Time&Acc&Time&\\
    \midrule
    EL2N w/o KD    & 59.71& 10.6h &63.50 & 17.1h &65.19& 31.3h &66.38& 35.0h  &67.90& 42.9h &69.01& 48.0h\\
    Glister w/o KD & 62.41& 11.9h &65.18 & 22.7h &66.47& 34.5h &67.05& 41.7h &68.81& 50.3h &69.45& 56.9h\\
    \midrule
    Ours w/o KD    & 66.91& 9.7h &68.77 & 16.0h &69.25& 29.5h &69.66& 34.4h &69.85& 40.2h &70.03& 45.4h\\
    \bottomrule
  \end{tabular}}
\end{table}

\section{Coreset Transferability and Generalizability} \label{sec:overlap}
The theoretical analysis of the importance of each sample in QAT is model-specific, which means that data selected using our method is the optimal coreset for the current model. However, if the coreset discovered by a specific pre-trained model is applicable to other models, our method can be a potential solution for the model-agnostic coreset method. In addition, if there is a significant coreset overlap across different models, our method can even solve dataset distillation.

We first design experiments to verify the transferability of the coreset using our method. We use 2/32-bit ResNet-18 to generate a coreset on CIFAR-100 and apply it to MobileNet-V2. The results are shown in Tab.~\ref{tab:transfer}. As the ResNet-18 coreset performs better than the random subset on MobileNet-V2, our coreset has some extent of generalization ability to unseen network structures, but the effectiveness is still worse than the MobileNetV2 coreset. 

\begin{table}[h]
  \caption{Comparision of Top-1 Accuracy with different coreset of MobileNet-V2 on CIFAR-100.}
  \label{tab:transfer}
  \centering
  \begin{tabular}{lcccccc}
    \toprule
    Method & Coreset Generation Model  & 10\% &20\% &30\% &40\% &50\%\\
    \midrule
    Random  &   -  & 62.25$\pm$0.71 &64.07$\pm$0.47 &65.22$\pm$0.41 &65.55$\pm$0.80 & 66.24$\pm$0.55   \\
    Ours&  MobileNetV2 & 63.37$\pm$0.55 & 65.91$\pm$0.40 & 66.41$\pm$0.31&66.86$\pm$0.72&67.19$\pm$0.65 \\
    Ours& ResNet-18 &62.94$\pm$0.45 & 64.18$\pm$0.73 & 65.70$\pm$0.40 &65.90$\pm$0.51 & 66.86$\pm$0.42\\
    \bottomrule
  \end{tabular}
\end{table}

We then design experiments to verify the overlap between the coreset of different models. We analyze the 10\% fraction coreset selected using our method in the final epochs of 2/32-bit ResNet-18, ResNet-50, and MobileNet-V2. The percentage of the overlap data is shown in Tab.~\ref{tab:overlap}.

\begin{table}[h]
  \caption{Coreset overlap of different model pairs on CIFAR-100.}
  \label{tab:overlap}
  \centering
  \begin{tabular}{cc}
    \toprule
    Model Pair & Coreset Overlap Percentage  \\
    \midrule
    \{ResNet-18, MobileNetV2\} & 37.1\% \\
    \{ResNet-18, ResNet-50\}   & \textbf{77.3\%} \\
    \{ResNet-50, MobileNetV2\} & 29.0\% \\
    \bottomrule
  \end{tabular}
\end{table}

We empirically find out that there is some overlap between the coreset of different models, and the overlap is more significant when these two models have a similar structure (ResNet-18 and ResNet-50). We then further apply majority voting from the CIFAR-100 coreset of five models (ResNet-18, ResNet-34, ResNet-50, MobileNet-V2, MobileNet-V3) to generate a “general coreset”. This “general coreset” of 10\% data fraction is applied to a 2/32-bit quantized Vision Transformer ViT-B/16~\citep{dosovitskiy2020image}, where we only got a Top-1 accuracy of 72.3\%, which is lower than 10\% coreset using our method of 78.9\% and even lower than 10\% random coreset of 74.6\%. We conclude that our method is model-specific, which has some extent of generalization ability to unseen network structures but cannot be a general data distillation approach.

\end{document}